\documentclass[10pt,twocolumn,letterpaper]{article}

\usepackage[pagenumbers]{cvpr} 

\usepackage{microtype}

\renewcommand{\paragraph}[1]{\vspace{.5em}\noindent\textbf{#1.}}

\definecolor{cvprblue}{rgb}{0.21,0.49,0.74}
\usepackage[pagebackref,breaklinks,colorlinks,allcolors=cvprblue]{hyperref}

\usepackage[accsupp]{axessibility}  

\usepackage{bm} 
\usepackage{multirow} 
\usepackage{makecell} 
\usepackage[ruled,vlined, linesnumbered, noend]{algorithm2e} 
\usepackage[misc]{ifsym} 
\usepackage{dblfnote} 

\def\paperID{} 
\def\confName{CVPR}
\def\confYear{2026}

\title{Collaborative Multi-Mode Pruning for Vision-Language Models}

\author{
Zimeng Wu\textsuperscript{1,2},
Yunhong Wang\textsuperscript{1,2},
Donghao Wang\textsuperscript{1,2},
Jiaxin Chen\textsuperscript{1,2, \Letter}\\
\textsuperscript{1}State Key Laboratory of Virtual Reality Technology and Systems, Beihang University, Beijing, China\\
\textsuperscript{2}School of Computer Science and Engineering, Beihang University, Beijing, China\\
{\tt\small \{zimengwu, yhwang, wonderhow, jiaxinchen\}@buaa.edu.cn}
}

\makeatletter
\newcommand{\LetterFootnote}[1]{%
  \begingroup
    \renewcommand\thefootnote{}%
    \def\@makefntext##1{\noindent ##1}%
    \footnotetext{\textsuperscript{\Letter} #1}%
  \endgroup
}
\makeatother

\begin{document}
\maketitle

\LetterFootnote{Corresponding Author}

\begin{abstract}
Vision-Language Models (VLMs) have advanced rapidly within the unified Transformer architecture, yet their deployment on resource-constrained devices remains challenging due to high computational complexity. 
While pruning has emerged as an effective technique for compressing VLMs, existing approaches predominantly focus on a single mode by pruning either parameters or tokens, neglecting fully exploring the inherent redundancy in each mode, which leads to substantial performance degradation at high pruning ratios. To address the above limitations, we propose Collaborative Multi-Mode Pruning (CoMP), a novel framework tailored for VLMs by performing joint parameter and token pruning.
Specifically, we first design a Collaborative Importance Metric (CIM) that investigates the mutual interference between the coupled parameters and tokens. It incorporates distinct significance of tokens into the computation of parameter importance scores, while simultaneously mitigating the affect of pruned parameters on token importance scores.
Moreover, we develop a Multi-Mode Pruning Strategy (MPS) that decomposes the overall pruning process into a sequence of pruning stages, while in each stage we estimate the priory of different pruning modes based on their pruning cost and adaptively shift to the optimal one. Additionally, MPS integrates the historical cost and random exploration, in order to achieve a stable pruning process and avoid local optimum.
Extensive experiments across various vision-language tasks and models demonstrate that our method effectively promotes the performance under high pruning ratios by comparing to the state-of-the-art approaches.
The source code is available at \url{https://github.com/Wuzimeng/CoMP.git}.
\end{abstract}
  
\section{Introduction}
\label{sec:introduction}

Vision-Language Models (VLMs), which integrate diverse modal spaces within a unified Transformer architecture \cite{transformer}, have become a cornerstone of recent advances in foundation models \cite{multimodal_foundation_survey, clip, blip}. They have proven effective across a wide range of downstream tasks via fine-tuning \cite{mingfang}, including visual reasoning \cite{nlvr2}, image-text retrieval \cite{flickr30k}, image captioning \cite{nocaps}, etc.
Despite their promising performance, the substantial computational overhead of the Transformer architecture continues to pose significant challenges to the deployment of VLMs on resource-constrained devices \cite{multimodal_foundation_survey, mope, upop}.

\begin{figure}[t]
    \centering
    \includegraphics[width=\linewidth]{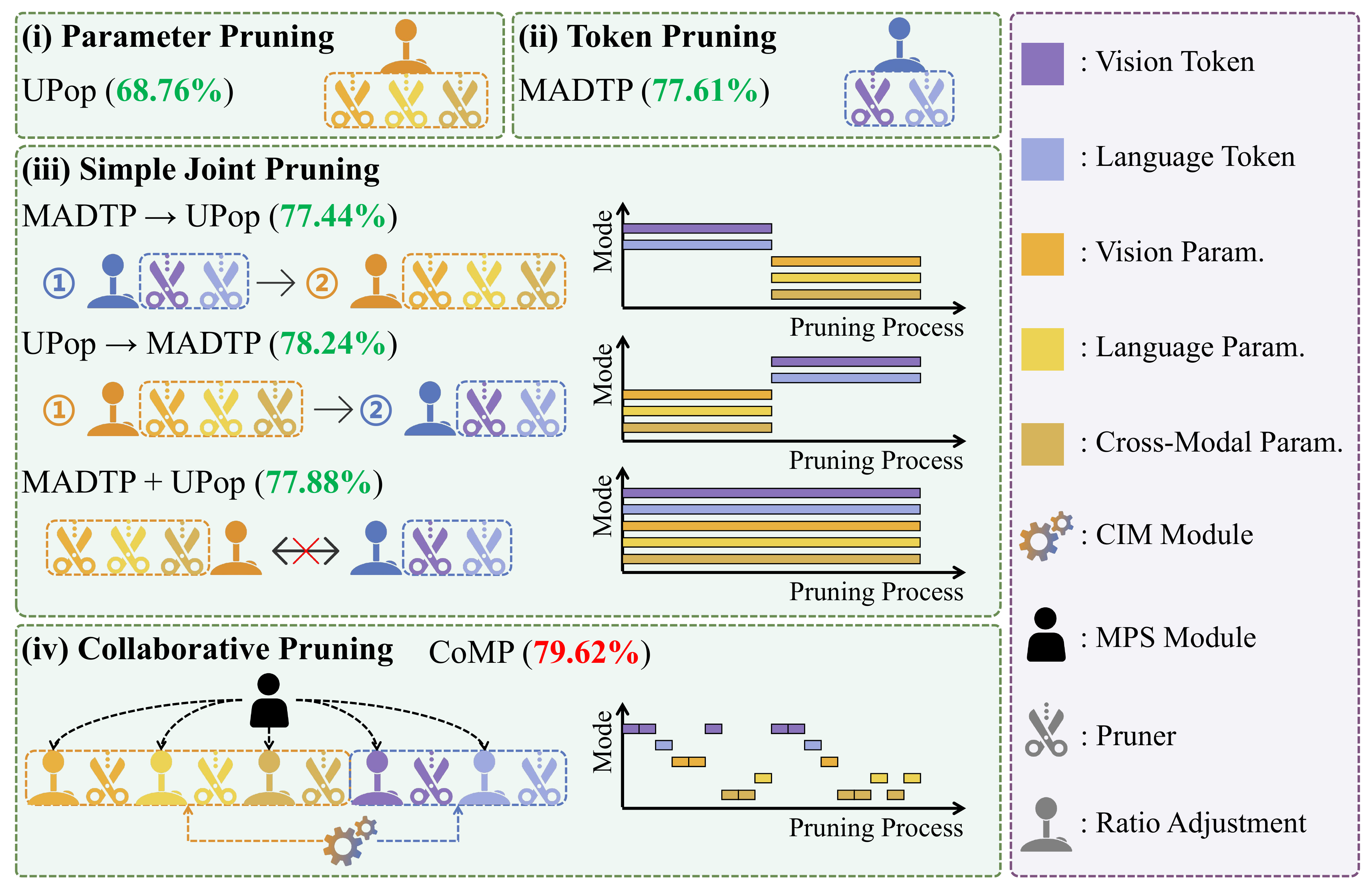}
    \caption{Illustration of different pruning modes for VLMs, with accuracy on NLVR2. For (i) parameter and (ii) token pruning, distinct modalities are simultaneously pruned under a unified ratio adjustment. For (iii) simple joint pruning, parameter and token pruning are conducted either sequentially or simultaneously without mitigating their inherent inconsistency. For (iv) our proposed CoMP, distinct pruning modes collaborate and only the optimal one is conducted at each stage in the progressive pruning process.}
    \label{fig:summary}
    \vspace{-10pt}
\end{figure}

In order to reduce the computational complexity, a variety of model compression techniques have been explored for VLMs \cite{mbq,tinyclip,jiayi}, among which model pruning stands out as a flexible and hardware-independent solution, garnering increasing attention in recent years \cite{prune_survey, transformer_compression_survey}. Generally, existing model pruning methods primarily concentrate on parameter pruning \cite{upop, ukmp} or token pruning \cite{madtp, crossget}. Given the inherent complexity of $O(N^2D+ND^2)$ for a standard Transformer block \cite{prance}, the former aims at diminishing the feature dimension $D$ by trimming network structures (\eg~learnable parameters) corresponding to insignificant channels, while the latter focuses on decreasing the sequence length $N$ by abandoning unimportant image or text tokens. Intuitively, the parameter pruning and token pruning play complementary roles as they reduce the redundancy residing in model structure and input data, respectively. However, early efforts typically perform single-mode pruning by either trimming parameters or tokens \cite{vit_slim, slimsam, ppt, upop, madtp}, thus resulting in suboptimal solution. This motivates our investigation: 
\textbf{Can joint pruning of both parameters and tokens achieve more effective compression of VLMs?}

\begin{figure}[!t]
  \centering
  \begin{subfigure}{0.48\linewidth}
    \captionsetup{skip=1pt,justification=centering}
    \centering
    \includegraphics[height=3cm, keepaspectratio]{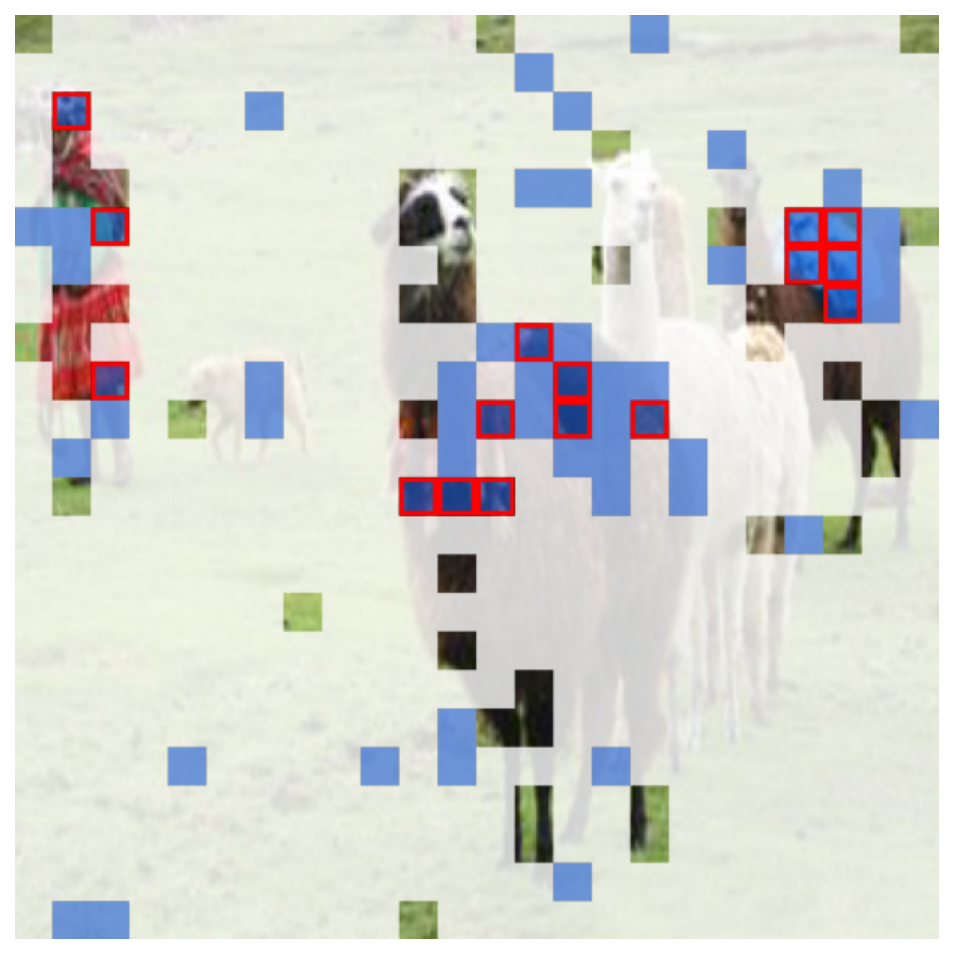}
    \caption{}
    \label{fig:imp_observation-a}
  \end{subfigure}
  \hfill
  \begin{subfigure}{0.48\linewidth}
    \captionsetup{skip=1pt,justification=centering}
    \centering
    \includegraphics[height=3cm, keepaspectratio]{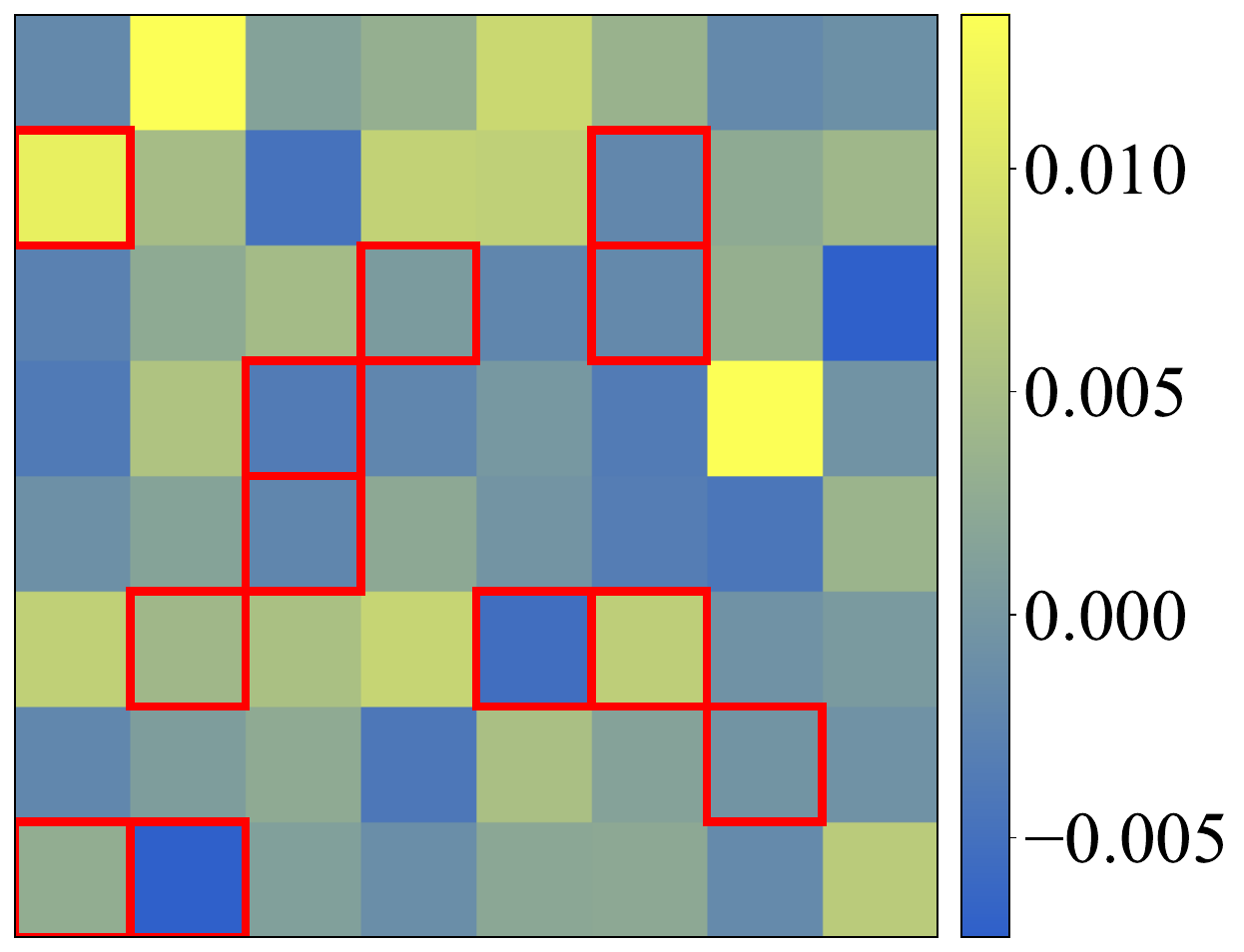}
    \caption{}
    \label{fig:imp_observation-b}
  \end{subfigure}
  \caption{(a) At $\mathit{layer}_{10}$ of BLIP's vision encoder, tokens contributing most to parameter importance (blue marks) and those with top token importance are only slightly (lower than 30\%) overlapped. (b) At $\mathit{layer}_2$ of BLIP's vision encoder, 75\% of the least important parameters (red boxes) still highly influence token importance.}
  \label{fig:imp_observation}
  \vspace{-10pt}
\end{figure}

As a preliminary exploration, we conduct empirical studies by establishing a straightforward multi-mode pruning baseline, directly trimming parameters and tokens jointly. As displayed in~\cref{fig:summary}, the simple combination fails to unleash the potential of joint pruning, only achieving comparable performance to single-mode pruning. 
We attribute this limitation to the following two core challenges: 
1) \emph{Intrinsic inconsistency between parameter and token importance metrics}. Existing methods typically employ separate importance metrics for parameters and tokens, and individually trim those with low importance scores. However, as shown in~\cref{fig:imp_observation-a}, the tokens contributing most in computation of parameter importance are vastly different from those with top scores measured by token importance metric, implying that unimportant tokens may dominate the estimation of parameter importance. Similarly, as illustrated in~\cref{fig:imp_observation-b}, the token importance metric depends on non-negligible amount of parameters with low importance scores that would be trimmed in parameter pruning. These inherent inconsistencies lead to suboptimal performance when simply performing joint pruning. 
2) \emph{Inflexible application of pruning modes}. Recent works typically perform pruning in a progressive way \cite{upop, madtp} by splitting the overall pruning process into multiple stages, in each of which different modes on distinct modalities are simultaneously conducted throughout the process in a fixed pattern until reaching the pruning ratio. 
However, as the model changes within distinct pruning stages, the optimal pruning mode, \ie~parameter or token pruning on the vision or language Transformer block, gradually shifts. As a consequence, applying different pruning modes in a fixed sequential order is prone to incur suboptimal solution, thus degrading the final performance under high pruning ratios.

\begin{figure*}[!t]
  \centering
  \begin{subfigure}{0.47\linewidth}
    \captionsetup{skip=1pt,justification=centering}
    \centering
    \includegraphics[height=4.5cm, keepaspectratio]{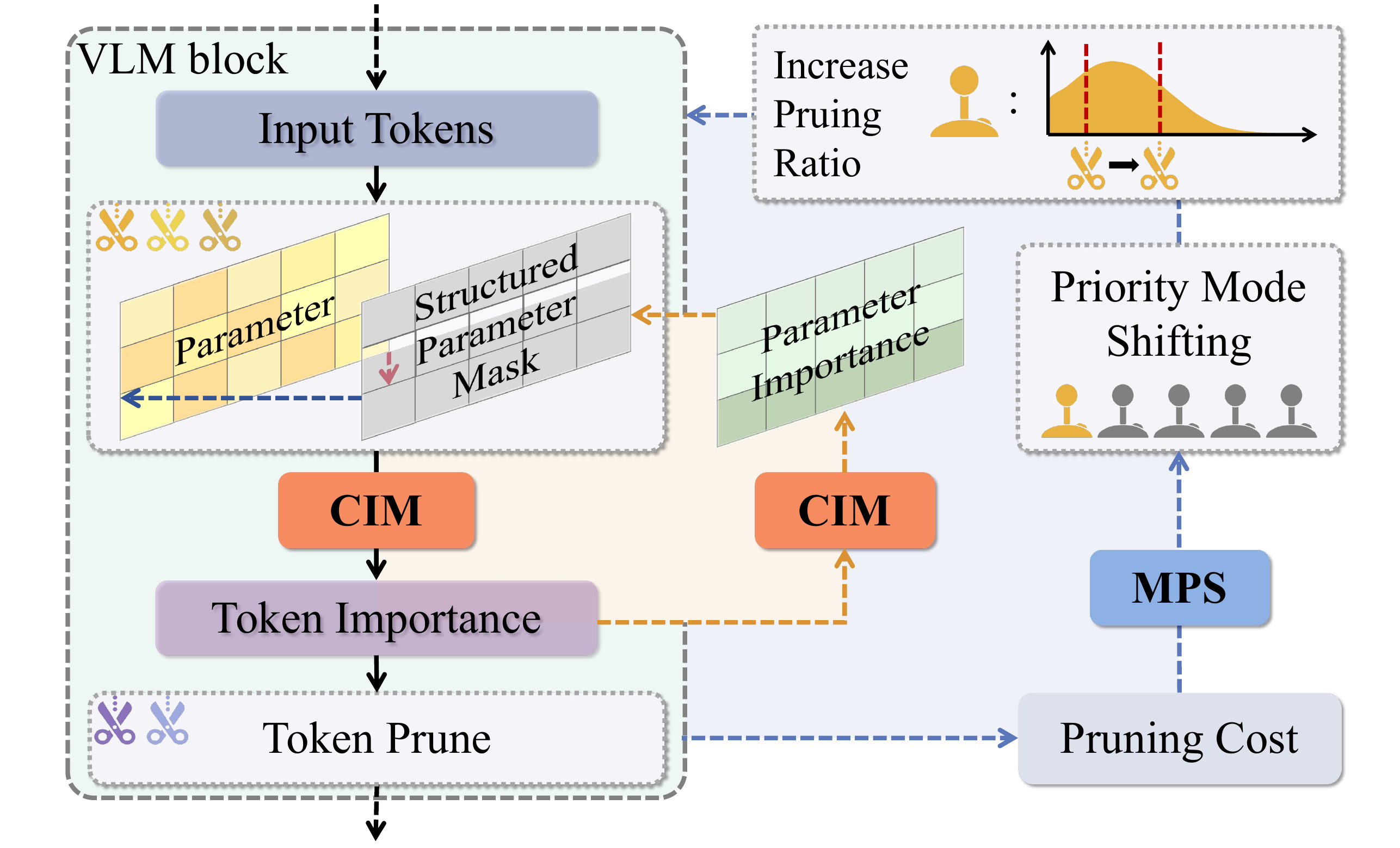}
    \caption{}
    \label{fig:framework-a}
  \end{subfigure}
  \hfill
  \begin{subfigure}{0.52\linewidth}
    \captionsetup{skip=1pt,justification=centering}
    \centering
    \includegraphics[height=4.5cm, keepaspectratio]{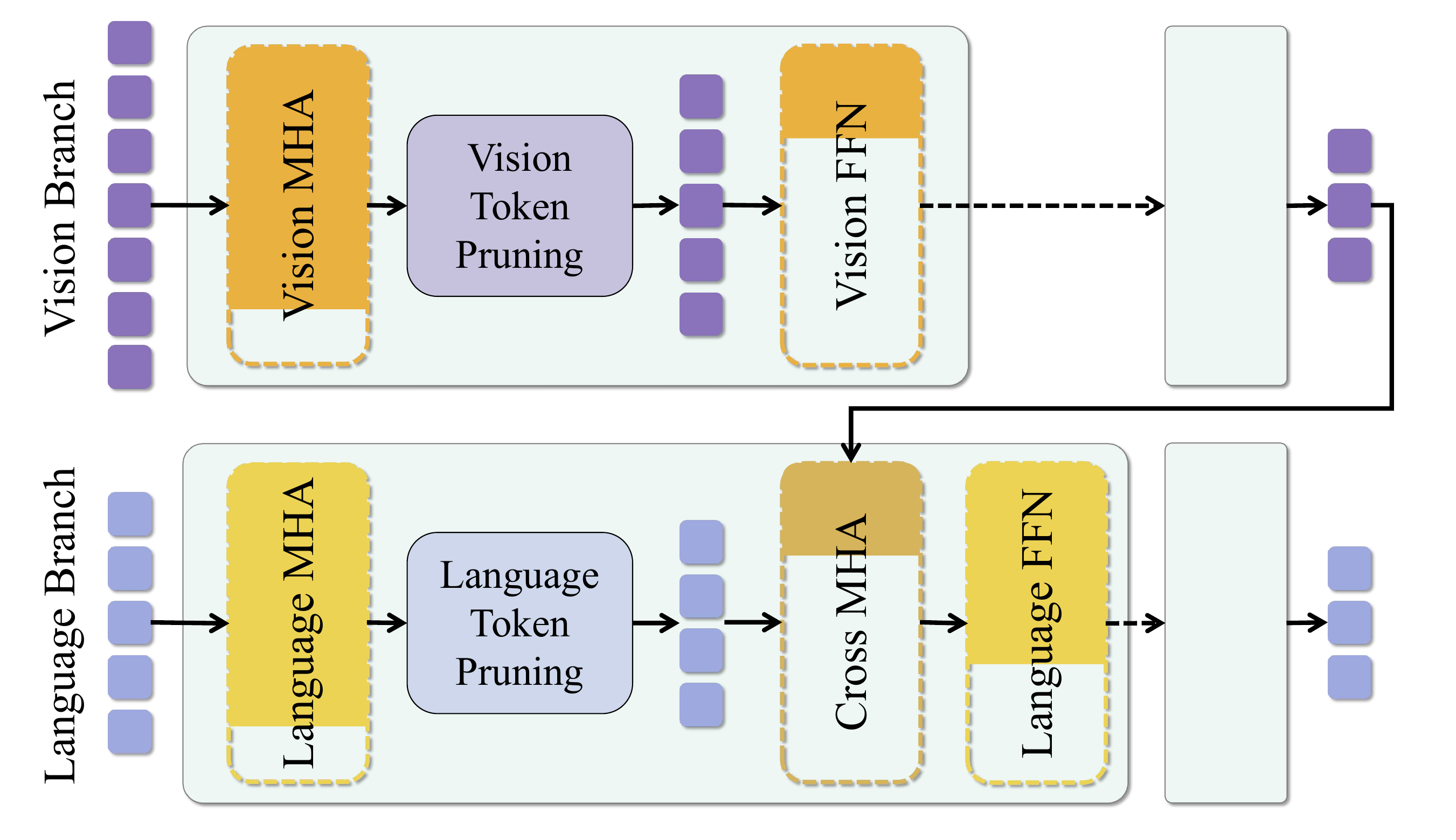}
    \caption{}
    \label{fig:framework-b}
  \end{subfigure}
  \caption{Framework overview of CoMP. (a) CoMP performs collaborative parameter and token pruning in nested loops. In the inner loop, input tokens are processed with partially masked parameters. The CIM module mitigates interference of progressive parameter pruning on token importance, and then suppresses the impact of redundant tokens for parameter importance. In the outer loop, the MPS module periodically selects the optimal pruning mode, whose corresponding threshold is adjusted to increase pruning ratio.
  (b) Given the full VLMs, CoMP compresses them by adaptively pruning parameters in different modalities, while enabling real-time token pruning during inference.}
  \label{fig:framework}
  \vspace{-5pt}
\end{figure*}

To address the above limitations, we propose a novel framework, namely Collaborative Multi-Mode Pruning (CoMP), for pruning VLMs. 
As shown in~\cref{fig:framework-a}, in order to deal with the inconsistency between distinct importance metrics, we first establish a Collaborative Importance Metric (CIM).
Concretely, CIM calculates parameter importance scores by incorporating token-weighted input norms, thereby mitigating distortions induced by tokens with low importance. Meanwhile, CIM refines token importance scores by transferring the parameter pruning mask to the attention weight matrix, effectively suppressing the influence of insignificant parameters. In addition, we develop an effective heuristic Multi-Mode Pruning Strategy (MPS) to identify the optimal mode in progressive pruning according to their respective pruning cost. MPS further leverages historical cost and incorporates random exploration to enhance stability and avoid convergence to locally optimal modes, ensuring a more robust and adaptive pruning process. 

In summary, our main contributions lie in three-fold:
\begin{itemize}
    \item We propose a novel pruning framework dubbed Collaborative Multi-Mode Pruning (CoMP) for VLMs, by exploring effective simultaneous pruning of parameters and tokens.
    \item We design a Collaborative Importance Metric (CIM) and a Multi-mode Pruning Strategy (MPS) to mitigate the inconsistency between distinct importance metrics and identify the optimal mode in progressive pruning, thus remarkably promoting the effectiveness of joint pruning.
    \item We extensively conduct experiments across diverse vision-language tasks and models, showing that our method outperforms state-of-the-art approaches, and achieves substantial performance gains, especially at high pruning ratios.
\end{itemize}

\section{Related Work}
\label{sec:related_work}
\subsection{Vision-Language Models}
Vision-Language Models have achieved remarkable progress by leveraging Transformer \cite{transformer} architectures for unified multimodal representation \cite{foundation_model_survey}.
Representative models like CLIP \cite{clip} and BLIP \cite{blip} serve as strong bases for task-specific VLMs by fine-tuning.
Mode recently, bridging vision encoders with Large Language Models (LLMs), \ie LLM-based VLMs, has emerged as a promising paradigm, such as LLaVA \cite{llava, llavaimproved}. Through visual instruction tuning, they achieve strong zero-shot image understanding\cite{vqav2,mmbench}.
Nevertheless, all these models require expensive computational overheard, underscoring the need for model compression.  

\subsection{Model Pruning}
\paragraph{Parameter Pruning}
Parameter pruning has proven effective in reducing model size and accelerating inference \cite{prune_quantize_survey, structured_prune_survey}.
Among various techniques, structured parameter pruning, which directly reduces the size of parameter matrices, is especially appealing due to its hardware-friendly nature \cite{samp, llmpruner}.
Typically, it involves a pipeline of parameter importance estimation, parameter removal and fine-tuning \cite{llmpruner, mope}.
To better estimate parameter contribution, various metrics have been developed, leveraging parameter magnitude \cite{isomorphic}, gradient \cite{group_fisher, miep}, input activations \cite{wanda}, etc.
However, these methods overlook token-wise contributions, which is one of the key issues our work aims to address.

\paragraph{Token Pruning}
Token pruning is devised for the quadratic complexity of Transformers \cite{dynamicvit}, and has demonstrated effectiveness in accelerating both vision \cite{dynamicvit, evit} and language models \cite{power_bert, spts}.
It performs token reduction in a cascaded manner \cite{spatten} usually by estimating their importance via attention mechanisms \cite{evit, zero_tprune, mstm, constraint_token_prune} or by learning auxiliary selection modules \cite{dynamicvit, a_vit, spvit}.
Another scheme for reducing token count is token merging \cite{tome, joint_token_prune_squeeze}, which, however, introduces extra computational overhead from computing token similarities.
While some work attempt to combine token pruning and merging \cite{joint_token_prune_squeeze, ppt}, they remain confined to data-level redundancy, neglecting the interference caused by redundant model parameters.

\paragraph{Model Pruning for VLMs}
There has been growing interest in pruning VLMs.
For structured parameter pruning, UPop \cite{upop} performs a unified progressive search for sub-networks, while MoPE-CLIP \cite{mope} designs a comprehensive scheme covering both pretraining and fine-tuning phases.
For token pruning, ELIP \cite{elip} prunes visual tokens under language supervision, Turbo \cite{turbo} merges tokens via information density, CrossGET \cite{crossget} and MADTP \cite{madtp} retain tokens for both modalities via cross-modal guidance. As for LLM-based VLMs, FastV \cite{fastv} exploits sparsity in visual attention to perform training-free token pruning. Subsequent works leverage the metrics of attention \cite{sparsevlm}, similarity \cite{visionzip, dart} and diversity \cite{divprune} to prune visual tokens at single \cite{divprune} or multiple \cite{pyramiddrop} layers. 
Supervised fine-tuning is also employed to recover the performance \cite{dart, visionzip, pyramiddrop}.

Most methods focus on either parameter or token pruning. While a few attempts have explored joint pruning \cite{turbo, corematching}, they adopt rigid schemes, yielding suboptimal performance. In contrast, we propose a more effective collaborative approach to enhance multi-mode compression for VLMs.

\section{Methodology}
\label{sec:methodology}
\subsection{Framework Overview}

We focus on the prevalent Transformer-based VLMs, which are typically composed of stacked Multi-Head Attention (MHA) and Feed-Forward Network (FFN) blocks.
Denote the block operation as $f$ and the inner non-linear operations (\eg~ attention and activation) as $\phi$, both types of the blocks can be formally expressed as $\bm{Z} = f(\bm{X}) = \phi(\bm{X}\bm{W}_{in})\bm{W}_{out} + \bm{X},$
where $\bm{X}, \bm{Z}\in \mathbb{R}^{N\times d}$ are the input and output with sequence length $N$ and embedding dimension $d$, $\bm{W}_{in}\in \mathbb{R}^{d\times D}$ and $\bm{W}_{out}\in \mathbb{R}^{D\times d}$ are the input and output projection matrices, respectively.

\begin{figure}[!t]
    \centering
    \begin{subfigure}[t]{0.8\linewidth}
        \captionsetup{skip=1pt,justification=centering}
        \centering
        \includegraphics[width=\linewidth]{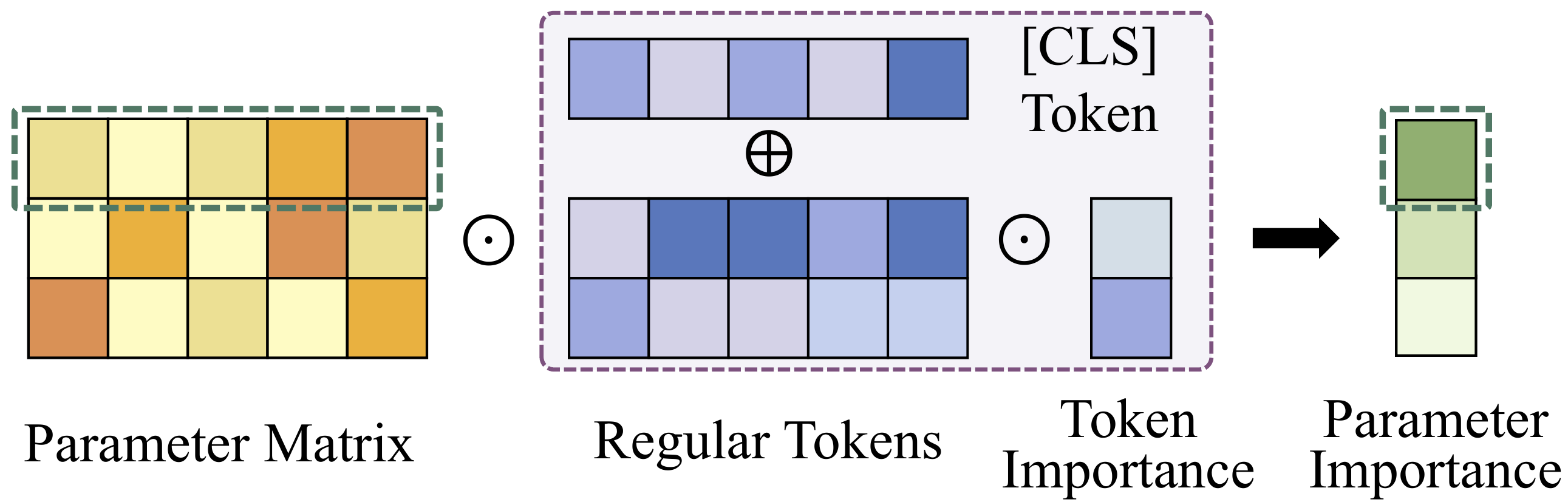}
        \caption{}
        \label{fig:cim_detail_a}
    \end{subfigure}
    \begin{subfigure}[t]{0.8\linewidth}
        \captionsetup{skip=1pt,justification=centering}
        \centering
        \includegraphics[width=\linewidth]{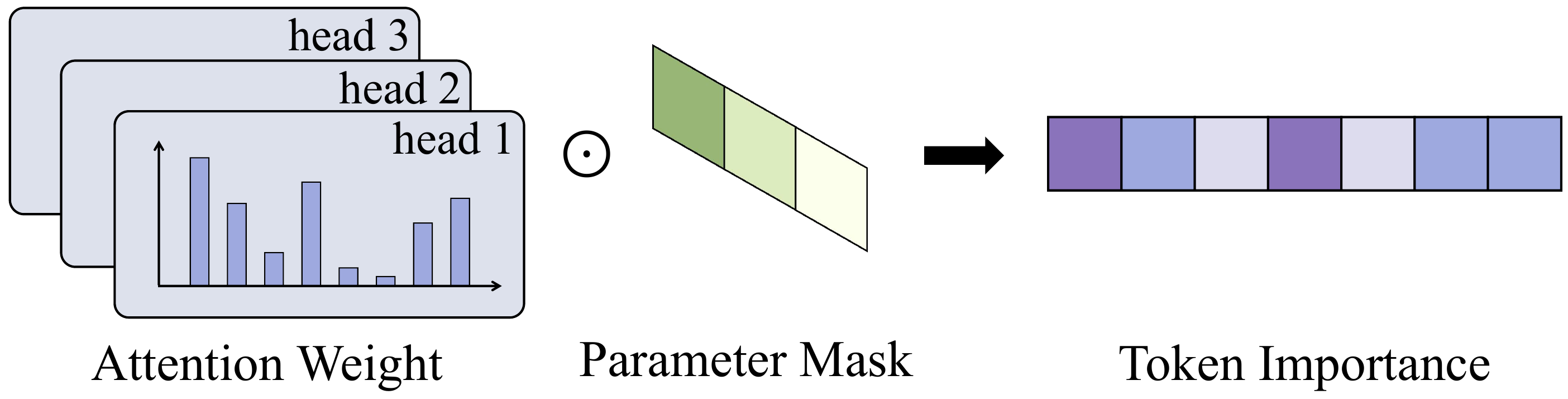}
        \caption{}
        \label{fig:cim_detail_b}
    \end{subfigure}
    \caption{Illustration of the CIM module. (a) adopts token-weighted input norm for parameter importance. (b) applies parameter pruning mask to the attention weight matrix for token importance.}
    \label{fig:cim_detail}
    \vspace{-5pt}
\end{figure}

Building on this, structured parameter pruning and token pruning aim at searching for binary masks $\bm{M}^p\in\{0, 1\}^{D}$ and $\bm{M}^t\in\{0,1\}^N$, which are applied via broadcasted Hadamard product as $\hat{\bm{W}}=\bm{W}\!\odot\!\bm{M}^p$ and $\hat{\bm{X}}=\bm{X}\!\odot\!\bm{M}^t$, indicating the components to be permanently discarded.

Our approach follows the general pruning paradigm, by heuristically estimating importance scores $\bm{S}^p\in \mathbb{R}^{D}$ and $\bm{S}^t\in \mathbb{R}^{N}$ for parameters and tokens, respectively, and building pruning masks according to the following formulation:
\begin{equation}
    \bm{M}^p=\mathbb{I}(\bm{S}^p > \theta^p),~~ \bm{M}^t=\mathbb{I}(\bm{S}^t > \theta^t),
    \label{eq:mask_w_thresh}
\end{equation}
where $\mathbb{I}$ is the indicator function, $\theta^p$ and $\theta^t$ denote optimizable thresholds governing the ratio for parameter and token pruning, respectively. 
For parameters, progressive pruning is typically implemented by gradually increasing $\theta^p$ and decreasing $\bm{M}^p$ from $1$ to $0$. The operation of each block can then be expressed as:
\begin{equation}
    \bm{Z} = \phi(\bm{X}\bm{W}_{in} \odot \bm{M}^p)\bm{W}_{out} + \bm{X}.
    \label{eq:mask_parameter}
\end{equation}
For tokens, progressive pruning is achieved solely by increasing $\theta^t$.
It performs a dynamic and cascade sequence reduction at layer $l$, formally as $\bm{X}^{l+1} = \bm{Z}^l \odot \bm{M}_t^l$.

As for the overall process, we combine the general paradigms of parameter and token pruning into a joint framework. 
To address the aforementioned challenges in multi-mode pruning for VLMs, we propose CoMP, which incorporates a Collaborative Importance Metric (CIM) and a Multi-Mode Pruning Strategy (MPS) for the optimization process as in ~\cref{fig:framework-a}, yielding a compressed network as in~\cref{fig:framework-b}.

\subsection{Collaborative Importance Metric}
\paragraph{Token-Weighted Parameter Importance Metric}
Data-driven importance metric is considered to better reflect the contribution of parameters to overall model performance \cite{mope, llmpruner}. 
However, due to the effect of the inherent LayerNorm operations in Transformer, existing metrics often fail to differentiate the contributions of individual tokens.
As a consequence, redundant tokens may mislead decisions on parameter pruning.
Therefore, we start from adopting a baseline metric computed solely from the forward pass, and further refine it with token-aware weighting.

\begin{figure}[t]
    \centering
    \includegraphics[width=\linewidth]{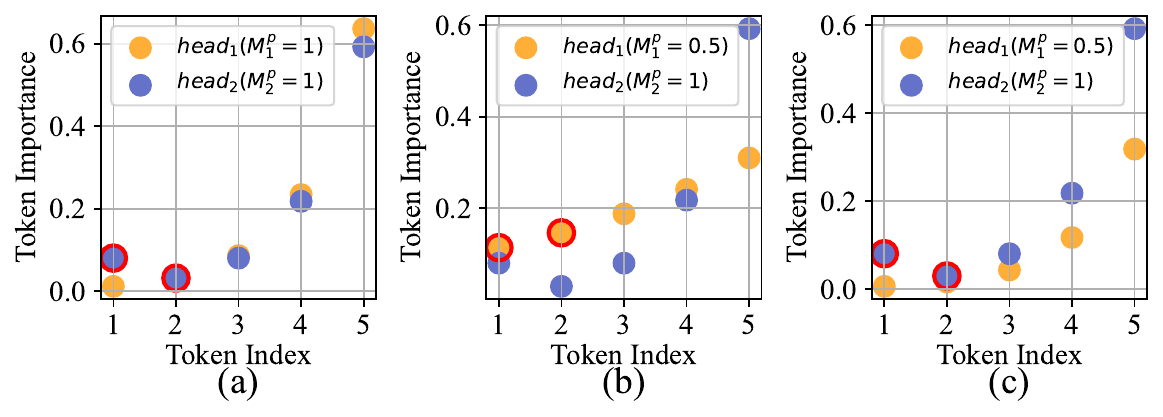}
    \caption{Illustration of interference between parameter pruning and token importance.
    (a) Without pruning, $token_1$ is more important than $token_2$.
    (b) Baseline pruning method masks redundant $head_1$ by~\cref{eq:mask_parameter}, flattens softmax, and distorts ranks of token importance.
    (c) By masking with~\cref{eq:mask_attention}, $head_1$ is gradually suppressed without disrupting correct ranks of token importance.}
    \label{fig:softmax_motivation}
    \vspace{-5pt}
\end{figure}

Following \cite{flap}, we first extend the Wanda metric \cite{wanda}, which is effective and efficient, to the structured pruning setting.\footnote{
For parallel inference, we follow common practice and perform head-wise pruning for MHA parameters, where channels within each head share the same mask and the averaged importance score. See \emph{Appendix} for details.}~
Formally, given the parameter matrix $\bm{W}\in \mathbb{R}^{D\times d}$ and its input $\bm{X}\in \mathbb{R}^{N\times D}$, the importance score for the $i$-th row of $\bm{W}$ is computed as:
\begin{equation}
    \bm{S}^p_{i,:} = \frac{1}{d}\sum_{j=1}^{d}|\bm{W}_{i,j}|\cdot ||\bm{X}_{:,i}||_2.
    \label{eq:wanda_imp}
\end{equation}

Building on this, we introduce an additional weighting mechanism using the token importance scores obtained from the token pruning process, and transforming $\bm{S}^p_{i,:}$ to $\bm{S}'^p_{i,:}$ as:
\begin{equation}
    \begin{split}
        \bm{S}'^p_{i,:} & = \frac{1}{d}\sum_{j=1}^{d}|\bm{W}_{i,j}|\cdot (\sum_{n=0}^{N}\omega_n\cdot \bm{X}_{n,i}^2)^\frac{1}{2}; \\
        \omega_0&=1; ~~\omega_n = \frac{\bm{S}^t_n}{\sum_{n=1}^{N}\bm{S}^t_n}, n>1.
    \end{split}
    \label{eq:our_weight_imp}
\end{equation}
Here, $n=0$ refers to the [CLS] token, and $n>0$ represents regular sequence tokens.
To emphasize the contribution of the global [CLS] and mitigate inter-layer discrepancies in token counts caused by token pruning, which may otherwise introduce norm scale biases, we normalize the token importance scores so that the total contribution of all regular tokens equals that of the [CLS] token. 
Additionally, for [CLS]-free models like LLaVA \cite{llavaimproved}, $\omega_0$ is omitted.

Furthermore, since tokens interact only through the attention mechanism, which also underlies token importance scores \cite{madtp, evit}, they can be considered independent between consecutive MHA blocks.  
Consequently, the token importance score obtained at layer $l$ is primarily determined by the computations of the MHA at the current layer and the FFN at the previous layer.
By denoting~\cref{eq:our_weight_imp} as $g$, we therefore distinguish the source of token importance, formally as:
\begin{equation}
\bm{S}'^{p,l} =
\begin{cases}
    g(\bm{W}^{l}, \bm{S}^{t,l+1}), & \text{if } \bm{W}^l \in \text{FFN}, \\
    g(\bm{W}^{l}, \bm{S}^{t,l}), & \text{if } \bm{W}^l \in \text{MHA}.
\end{cases}
\end{equation}

\paragraph{Self-corrected Token Importance Metric}
For token pruning, we adopt the prevalent attention-based baseline, defining the importance of the $i$-th token as:
\begin{equation}
        \bm{S}^t_i = \mathrm{Norm}(\sum_{n=1}^N\max_{h=1,2,...,H}\bm{A}_{h,n,i}),
    \label{eq:token_imp}
\end{equation}
where $\bm{A}\in\mathbb{R}^{H\times N\times N}$ is the attention matrix for $H$ heads. It is computed as $\mathrm{Softmax}(\bm{Q}\bm{K}^T/{\sqrt{d_k}})$, with $\bm{Q}, \bm{K}$ the query and key matrices, and $d_k\!=\!\frac{D}{H}$ the per-head dimension.

Redundancy across attention heads has been widely acknowledged \cite{ukmp, llmpruner}.
While pruning mask can already suppress their influence, existing pruning mechanisms as~\cref{eq:mask_parameter} may allow redundant heads to interfere token importance with a nearly uniform attention distribution due to the normalization effect of Softmax, as illustrated in~\cref{fig:softmax_motivation}.

To address this issue, we propose applying the parameter pruning mask directly to $\bm{A}$ for correction, as shown in~\cref{eq:mask_attention}. 
$\hat{\bm{M}}^p\in\mathbb{R}^H$ is formed by selecting every $\frac{D}{H}$-th value from $\bm{M}^p$ to match the size of $\bm{A}$.~
The resulting $\hat{\bm{A}}$ is subsequently used for~\cref{eq:token_imp}.
\begin{equation}
    \hat{\bm{A}} = \bm{A}\odot \hat{\bm{M}}^p
    \label{eq:mask_attention}.
\end{equation}

\subsection{Multi-Mode Pruning Strategy}
\paragraph{Pruning Cost-Aware Mode Shifting}
Based on modality and redundant types, we first define five distinct pruning modes: $\mathcal{B}\!=\!\{B^p_v, B^p_l, B^p_c, B^t_v, B^t_l\}$, representing pruning of vision parameter, language parameter, cross-modal parameter, vision token and language token.
Each mode is associated with an optimizable threshold in $\Theta\!=\!\{\theta^p_v, \theta^p_l, \theta^p_c, \theta^t_v, \theta^t_l\}$ to adjust its pruning ratio. The progressive pruning process consists of selecting a pruning mode $\mathcal{B}_m$, increasing the corresponding pruning threshold $\Theta_m$, followed by updating model parameters via training.

However, due to the difficulty of analytically modeling the precise future impact of each mode at a given pruning stage, we maintain a cost estimate $\mathcal{R}\!=\!\{r^p_v, r^p_l, r^p_c, r^t_v, r^t_l\}$ to guide the selection of priority modes. 
After conducting pruning mode $\mathcal{B}_m$, we update its cost and determine the next mode by the following:
\begin{equation}
    r=\frac{\Delta \mathit{val\_acc}}{\Delta \mathit{FLOPs}}, ~~\mathcal{R}_m^{\text{cur}}\leftarrow r, ~~m\leftarrow \arg \min(\mathcal{R}).
    \label{eq:prune_cost}
\end{equation}
$r$ quantifies the sensitivity of model to the executed pruning mode by measuring the validation accuracy change per unit FLOPs, and is used to update the maintained cost $\mathcal{R}_m^{\text{cur}}$.
Finally, the pruning mode with the lowest cost is selected as the target for the next pruning stage.

\paragraph{Priority Refinement with Exploration and Historical Information}
While effective, the fully greedy strategy described above may lead to prolonged focus on a single mode, thus increasing the risk of convergence to suboptimal local minima with over-pruning.
To better balance the trade-off between minimizing cumulative pruning cost and maintaining diversity across pruning modes, we enhance the mode-shifting mechanism with a mixed strategy. Specifically, at each pruning stage, a pruning mode is randomly selected with probability $\rho$, instead of strictly choosing the one with the lowest pruning cost.

Moreover, to guide this random selection and promote diversity, we incorporate historical information.
Concretely, we maintain a timestamp $\mathcal{T}\!=\!\{T^p_v, T^p_l, T^p_c, T^t_v, T^t_l\}$ to record the most recent execution stage of each pruning mode.
Denoting the interval since the last execution of $\mathcal{B}_m$ as $I_m(\ge 1)$, the probability $\rho_m$ of choosing $\mathcal{B}_m$ at stage $T$ is determined by using an interval-weighted Softmax:
\begin{equation}
    I_m = T-\mathcal{T}_m, ~~\rho_m = \mathrm{Softmax}(I_m/\tau),
    \label{eq:rho_m}
\end{equation}
which favors modes that have not been performed recently.

We also incorporate this time interval into the pruning cost to stabilize pruning process.
Formally as~\cref{eq:ema_cost}, we maintain an Exponential Moving Average (EMA) of the observed single-stage cost for each mode:
\begin{equation}
\begin{aligned}
    \mathcal{R}_m^{\text{cur}} \leftarrow \lambda \mathcal{R}_m^{\text{pre}} + (1 - \lambda) r,\quad~~ \\
    \lambda = \max\left( \lambda_0 - \frac{\lambda_0}{I_{\text{max}}}(I_m\!-\!1), 0\right),
\end{aligned}
\label{eq:ema_cost}
\end{equation}
where $\mathcal{R}_m^{\text{cur}}$ and $\mathcal{R}_m^{\text{pre}}$ denote the updated and previous pruning cost values, respectively. $\lambda$ is a decay factor balancing historical information with current feedback.
It decreases linearly from $\lambda_0$ to $0$, ensuring that only the pruning cost in the most recent $I_{\text{max}}$ stages contributes to the current mode shifting decision.

\section{Experimental Results and Analysis}

\begin{table}[!t]
    \centering
    \caption{Comparison of Dev./Test Acc. (\%) and GFLOPs by various pruning methods for BLIP on NLVR2 with distinct pruning ratios for the visual reasoning task. `P', `T', `J' and `C' denote parameter pruning, token pruning, joint pruning and collaborative pruning, respectively. `SJP' stands for the simple joint pruning baseline. The best results are highlighted in \textbf{bold}.}
    \label{tab:sota_table}
    \footnotesize
    \begin{tabular}{@{\hskip 5pt} c @{\hskip 5pt}|@{\hskip 5pt} c @{\hskip 5pt}|@{\hskip 5pt} c @{\hskip 5pt}| c c |@{\hskip 5pt} c @{\hskip 5pt}}
        \toprule[1pt]
        \multirow{2}{*}{\makecell[c]{Method}} & \multirow{2}{*}{\makecell{Pruning \\ Mode}} & \multirow{2}{*}{\makecell{Pruning \\ Ratio}} & \multirow{2}{*}{\makecell{Dev. \\ Acc.}} & \multirow{2}{*}{\makecell{Test \\ Acc.}} & \multirow{2}{*}{GFLOPs} \\
         & & & & & \\
        \midrule
        \makecell[l]{Uncompressed}            & /                  & /            & 82.48          & 83.08          & 132.54 \\
        \midrule
        \makecell[l]{M-Pruning \cite{upop}}   & P                  & 0.5          & 75.74          & 76.44          & 66.88 \\
        \makecell[l]{UPop \cite{upop}}        & P                  & 0.5          & 76.89          & 77.61          & 65.29\\ 
        \makecell[l]{STP \cite{madtp}}        & T                  & 0.5          & 78.08          & 77.61          & 68.31\\
        \makecell[l]{MADTP \cite{madtp}}      & T                  & 0.5          & \textbf{81.75} & 82.35          & 66.36\\
        \makecell[l]{\textbf{CoMP (Ours)}}    & C                  & 0.5          & 81.67          & \textbf{82.61} & 66.74\\
        \midrule
        \makecell[l]{UPop \cite{upop}}        & P                  & 0.6          & 72.85          & 73.55          & 50.35\\
        \makecell[l]{MADTP \cite{madtp}}      & T                  & 0.6          & 81.07          & \textbf{82.10} & 52.73\\
        \makecell[l]{Turbo \cite{turbo}}      & T                  & 0.6          & 80.50           & 81.50           & 62.20\\
        \makecell[l]{SJP$_{U\rightarrow T}$ \cite{turbo}} & J      & 0.6          & 79.00           & 79.80           & 54.70\\
        \makecell[l]{\textbf{CoMP (Ours)}}    & C                  & 0.6          & \textbf{81.28} & 81.99          & 52.66\\
        \midrule
        \makecell[l]{UPop \cite{upop}}        & P                  & 0.7          & 68.71          & 68.76          & 39.93\\
        \makecell[l]{MADTP \cite{madtp}}      & T                  & 0.7          & 80.34          & 80.78          & 39.63\\
        \makecell[l]{\textbf{CoMP (Ours)}}    & C                  & 0.7          & \textbf{80.95} & \textbf{81.37} & 39.72\\
        \midrule
        \makecell[l]{UPop \cite{upop}}        & P                  & 0.8          & 67.80          & 67.49          & 30.78\\
        \makecell[l]{MADTP \cite{madtp}}      & T                  & 0.8          & 77.21          & 77.61          & 26.46\\
        \makecell[l]{SJP$_{U\rightarrow M}$}  & J                  & 0.8          & 77.61          & 78.27          & 26.54\\
        \makecell[l]{SJP$_{M\rightarrow U}$}  & J                  & 0.8          & 77.23          & 77.44          & 26.61\\
        \makecell[l]{SJP$_{M + U}$}           & J                  & 0.8          & 76.69          & 77.88          & 26.39\\
        \makecell[l]{\textbf{CoMP (Ours)}}    & C                  & 0.8          & \textbf{79.23} & \textbf{79.62} & 25.97\\
        \midrule
        \makecell[l]{UPop \cite{upop}}        & P                  & 0.85         & 62.09          & 62.10          & 20.01\\
        \makecell[l]{MADTP \cite{madtp}}      & T                  & 0.85         & 71.50          & 72.57          & 20.57\\
        \makecell[l]{\textbf{CoMP (Ours)}}    & C                  & 0.85         & \textbf{75.81} & \textbf{76.08} & 20.26\\
        \bottomrule[1pt]
    \end{tabular}
\end{table}

\subsection{Experimental Settings}
\paragraph{Models}
By following prior works \cite{upop, madtp} we mainly evaluate our method on BLIP \cite{blip} and CLIP \cite{clip} that are initially fine-tuned by \cite{upop} for various downstream tasks. 
We also extend our method to LLaVA-v1.5-7B \cite{llavaimproved} that represents the prevalent LLM-based VLMs. 

\paragraph{Datasets and Evaluation Metrics}
\label{sec:experiment_dataset}
For task-specific VLMs, we evaluate CoMP on four datasets. Concretely, we report accuracy on the development (Dev. Acc.) and test (Test Acc.) sets of NLVR2 \cite{nlvr2} for the visual reasoning task. For image-text retrieval, we report the top-1/-5 recall, denoted by R@1/5, on COCO \cite{coco} and Flickr30K \cite{flickr30k}. CIDEr \cite{cider} and SPICE \cite{spice} on COCO \cite{coco} are presented for image captioning, and accuracy on the development (Test-dev) and standard (Test-std) sets of VQAv2 \cite{vqav2} are reported for visual question answering.
As for LLaVA, pruning is conducted during training on the official 665K instruction data \cite{llavaimproved}. We evaluate methods on six widely used image understanding benchmarks, by reporting the sum of perception and cognition scores for MME \cite{mme}, accuracy for MMBench (denoted as MMB) \cite{mmbench}, GQA \cite{gqa}, TextVQA \cite{textvqa} and VQAv2 \cite{vqav2}, and F1 score for POPE \cite{pope}.
To demonstrate computational efficiency, we report the Floating-Point Operations (\ie GFLOPs or TFLOPs) per sample during inference.
More detailed descriptions are provided in the \emph{Appendix}.

\begin{table*}[!t]
    \centering
    \caption{(\textbf{Top}) Comparison of R@1/5 (\%) and GFLOPs for BLIP/CLIP on Flickr30K and COCO for the image-text retrieval task. (\textbf{Bottom}) Comparison of CIDEr and SPICE on COCO for the image captioning task and Test-dev/std accuracy (\%) on VQAv2 for the visual question answering task as well as GFLOPs, based on BLIP. `*' indicates GFLOPs during inference by our re-implementation, which were not reported in the original works. The best results are highlighted in \textbf{bold}.}
    \label{tab:other_tasks_table}
    \footnotesize
    \begin{tabular}{c| c| c| c| c  c | c  c | c | c  c | c  c | c }
        \toprule[1pt]
        \multirow{3}{*}{Model} & \multirow{3}{*}{Method} & \multirow{3}{*}{\makecell{Pruning \\ Mode}} & \multirow{3}{*}{\makecell{Pruning \\ Ratio}} & \multicolumn{5}{c|}{Flickr30K (1K test set)} & \multicolumn{5}{c}{COCO (5K test set)} \\
        \cline{5-9} \cline{10-14}
         & & & & \multicolumn{2}{c|}{I$\rightarrow$T} & \multicolumn{2}{c|}{T$\rightarrow$I} & \multirow{2}{*}{GFLOPs} & \multicolumn{2}{c|}{I$\rightarrow$T} & \multicolumn{2}{c|}{T$\rightarrow$I} & \multirow{2}{*}{GFLOPs} \\
         & & & & R@1 & R@5 & R@1 & R@5 &  & R@1 & R@5 & R@1 & R@5 & \\
         \midrule
        \multirow{4}{*}{BLIP} & \makecell[l]{Uncompressed}         & / & /   & 96.8 & 99.9 & 86.9 & 97.3 & 91.65 & 81.9 & 95.4 & 64.3 & 85.7 & 91.65 \\
                              & \makecell[l]{UPop\textsuperscript{*} \cite{upop}}    & P & 0.7 & 88.4 & 98.9 & 76.9 & 94.5 & 35.21 & 73.0 & 91.9 & 55.5 & 81.1 & 34.33 \\ 
                              & \makecell[l]{MADTP\textsuperscript{*} \cite{madtp}}  & T & 0.7 & 92.6 & 99.1 & 79.0 & 94.6 & 30.96 & 73.9 & 91.5 & 56.0 & 80.6 & 30.69 \\ 
                              & \makecell[l]{\textbf{CoMP (Ours)}} & C & 0.7 & \textbf{94.4} & \textbf{99.6} & \textbf{80.1} & \textbf{95.1} & 30.47 & \textbf{76.2} & \textbf{92.4} & \textbf{57.0} & \textbf{81.1} & 30.08 \\ 
        \midrule
        \multirow{4}{*}{CLIP} & \makecell[l]{Uncompressed}         & / & /    & 96.8 & 100.0 & 86.6 & 97.8 & 197.8 & 71.5 & 90.8 & 56.8 & 80.7 & 197.8 \\
                              & \makecell[l]{UPop\textsuperscript{*} \cite{upop}}     & P & 0.75 & 82.9 & 95.7  & 67.3 & 89.5 & 51.3 & 56.1 & 82.4 & 41.1 & 71.0 & 57.9 \\ 
                              & \makecell[l]{MADTP\textsuperscript{*} \cite{madtp}}   & T & 0.75 & 88.4 & 97.3  & 76.9 & 94.2 & 49.7 & 66.2 & 88.4 & 49.9 & 76.3 & 46.2 \\ 
                              & \makecell[l]{\textbf{CoMP (Ours)}} & C & 0.75 & \textbf{88.6} & \textbf{98.6} & \textbf{78.5} & \textbf{94.6} & 44.7 & \textbf{68.7} & \textbf{88.9} & \textbf{51.7} & \textbf{77.6} & 45.1 \\ 
        \cmidrule[1pt]{1-14}
        \multirow{2}{*}{Model} & \multirow{2}{*}{Method} & \multirow{2}{*}{\makecell{Pruning \\Mode}} & \multirow{2}{*}{\makecell{Pruning \\ Ratio}} & \multicolumn{5}{c|}{Image Caption (COCO)} & \multicolumn{5}{c}{VQA (VQAv2)} \\
        \cline{5-9} \cline{10-14}
         & & & & \multicolumn{2}{c|}{CIDEr} & \multicolumn{2}{c|}{SPICE} & GFLOPs & \multicolumn{2}{c|}{Test-dev} & \multicolumn{2}{c|}{Test-std} & GFLOPs \\
        \midrule
        \multirow{4}{*}{BLIP} & \makecell[l]{Uncompressed} & / & / & \multicolumn{2}{c|}{133.3} & \multicolumn{2}{c|}{23.8} & 65.7 & \multicolumn{2}{c|}{77.4} & \multicolumn{2}{c|}{77.5} & 186.1 \\
         & \makecell[l]{UPop \cite{upop}} & P & 0.7 & \multicolumn{2}{c|}{117.4} & \multicolumn{2}{c|}{21.7} & 22.2 & \multicolumn{2}{c|}{74.5} & \multicolumn{2}{c|}{74.6} & 62.3 \\
         & \makecell[l]{MADTP \cite{madtp}} & T & 0.7 & \multicolumn{2}{c|}{120.1} & \multicolumn{2}{c|}{22.0} & 22.1 & \multicolumn{2}{c|}{76.3} & \multicolumn{2}{c|}{76.2} & 61.6 \\
         & \makecell[l]{\textbf{CoMP (Ours)}} & C & 0.7 & \multicolumn{2}{c|}{\textbf{126.8}} & \multicolumn{2}{c|}{\textbf{23.1}} & 21.2  & \multicolumn{2}{c|}{\textbf{76.5}} & \multicolumn{2}{c|}{\textbf{76.5}} & 59.7 \\
        \bottomrule[1pt]
    \end{tabular}
\end{table*}

\paragraph{Implementation Details}
\label{sec:experiment_implementation}
We employ the framework in UPop \cite{upop} for parameter pruning. And we perform token pruning based on MADTP \cite{madtp} for BLIP/CLIP, and PDrop \cite{pyramiddrop} for LLaVA, respectively.
Notably, these frameworks serve as single-mode pruning baselines, while our CoMP complements multi-mode collaboration.
The thresholds $\theta^p$ and $\theta^t$ that control the pruning ratios follow the settings used in baseline frameworks, with a slight extension to support multiple modes for finer-grained collaborative optimization.
In MPS, we set $\rho=0.2$ for random exploration and fix $\tau=5$ in~\cref{eq:rho_m}.
In~\cref{eq:ema_cost}, $\lambda_0$ and $I_{\text{max}}$ are fixed to 0.4 and 5, respectively.
Following typical practice, CoMP first progressively prunes the model to the target FLOPs, then the pruning configuration is fixed and the pruned model is fine-tuned to recover performance.
All experiments are conducted on 2 NVIDIA A800 GPUs, unless stated otherwise.
Additional training details and hyperparameter ablations are depicted in the \emph{Appendix}.

\subsection{Main Results on Vision-Language Tasks}
\label{sec:main_results}
We compare our proposed CoMP with the following state-of-the-art pruning approaches for task-specific VLMs: 1) parameter pruning including UPop and Mask-based Pruning (\ie~M-Pruning) as in \cite{upop}; 2) token pruning including MADTP, STP as in \cite{madtp} and Turbo \cite{turbo}. We also compare to the simple joint pruning baselines as shown in~\cref{fig:summary}.

\paragraph{On Visual Reasoning Task}
We evaluate the effectiveness of CoMP on visual reasoning task by pruning the BLIP \cite{blip} model trained on NLVR2 \cite{nlvr2}, with pruning ratios ranging from 0.5 to 0.85.
As shown in~\cref{tab:sota_table}, CoMP performs comparably at medium pruning ratios (\ie~$\le$ 0.6), and consistently outperforms the compared approaches at higher pruning ratios (\ie~$\ge$ 0.7) where single-mode pruning is prone to overly trimmed tokens or parameters. For instance, it achieves a 3.51\% improvement in test accuracy at a pruning ratio of 0.85. 
The results also suggest that the early-stage redundancy in VLMs mainly stems from tokens, where CoMP adaptively shifts to the token pruning mode, yielding behavior similar to the token pruning baseline MADTP. 
However, as the pruning ratio increases, the redundancies in both modes become more comparable, leading to increasingly significant mutual interference. At this stage, CoMP facilitates more effective collaboration, thereby exhibiting more substantial advantages.
In addition, Turbo \cite{turbo} introduces a simple joint pruning variant by combining with UPop (\ie~SJP$_{U \rightarrow T}$), yet CoMP still surpasses it by 2.19\% at a pruning ratio of 0.6, further highlighting the efficacy of our collaborative design.

\paragraph{On Image-Text Retrieval Task}
We compare CoMP with UPop \cite{upop} and MADTP \cite{madtp} on image-text retrieval, by pruning the BLIP \cite{blip} and CLIP \cite{clip} models fine-tuned on COCO \cite{coco} and Flickr30K \cite{flickr30k}.
As displayed in~\cref{tab:other_tasks_table} (Top), at pruning ratios above 0.7, CoMP consistently improves performance, increasing recall@1 on COCO by 2.3\% and 2.5\% for BLIP and CLIP, respectively.
These results also demonstrate the broad applicability of our method to diverse VLM architectures.

\paragraph{On Image Captioning Task}
To further evaluate the generalization capability of our method, we conduct experiments on the image captioning task at a pruning ratio of 0.7, of which the performance is typically sensitive to pruning \cite{ukmp}.
As shown in~\cref{tab:other_tasks_table} (Bottom), CoMP outperforms compared approaches, surpassing the second best MADTP \cite{madtp} by 6.7 and 1.1 in CIDEr and SPICE scores, respectively.

\begin{table*}[!t]
    \centering
    \caption{Comparison of performance and TFLOPs by various pruning methods on LLaVA-v1.5-7B with distinct pruning ratios on $6$ commonly-used benchmarks. The `Average' column summarizes the average score across all tasks. `$\dagger$' indicates that supervised fine-tuning is involved. The best results are highlighted in \textbf{bold} and the second-best results are \underline{underlined}.}
    \label{tab:llava_comparison}
    \footnotesize
    \begin{tabular}{@{\hskip 8pt}c@{\hskip 8pt}|@{\hskip 8pt} c @{\hskip 8pt}|@{\hskip 8pt} c @{\hskip 8pt}|@{\hskip 8pt} c@{\hskip 8pt}|@{\hskip 8pt} c@{\hskip 8pt}|@{\hskip 8pt} c@{\hskip 8pt}|@{\hskip 8pt} c@{\hskip 8pt}|@{\hskip 8pt} c@{\hskip 8pt}|@{\hskip 8pt} c@{\hskip 8pt}|@{\hskip 8pt} c@{\hskip 8pt}|@{\hskip 8pt} c @{\hskip 8pt}}
        \toprule[1pt]
        \multirow{2}{*}{Method} & \multirow{2}{*}{\makecell{Pruning \\ Mode}} & \multirow{2}{*}{\makecell{Pruning \\ Ratio}} & \multirow{2}{*}{MME} & \multirow{2}{*}{MMB} & \multirow{2}{*}{GQA} & \multirow{2}{*}{TextVQA} & \multirow{2}{*}{VQAv2} & \multirow{2}{*}{POPE} & \multirow{2}{*}{Average} & \multirow{2}{*}{TFLOPs} \\
         & & & & & & & & & & \\
        \midrule
        \makecell[l]{Uncommpressed}                  & / & /    & 1862 & 64.7 & 61.9 & 58.2 & 78.5 & 85.9 & 69.28 & 5.63 \\
        \midrule
        \makecell[l]{FastV \cite{fastv}}             & T & 0.46 & 1612 & 61.2 & 52.7 & 52.5 & 67.1 & 64.8 & 59.31 & 3.18 \\
        \makecell[l]{SparseVLM \cite{sparsevlm}}     & T & 0.46 & 1787 & 64.1 & 59.5 & \textbf{57.8} & 75.6 & 85.3 & 67.69 & 3.01 \\
        \makecell[l]{PDrop \cite{pyramiddrop}}       & T & 0.46 & 1797 & 63.3 & 57.3 & 56.5 & 75.1 & 82.3 & 66.45 & 3.09 \\
        \makecell[l]{VisionZip \cite{visionzip}}     & T & 0.46 & 1783 & 63.0 & 59.3 & 57.3 & 76.8 & 85.3 & 67.56 & 3.01 \\
        \makecell[l]{DART \cite{dart}}               & T & 0.46 & \textbf{1856} & 63.6 & 60.0 & 57.4 & 76.7 & 82.8 & 67.80 & 3.18 \\
        \makecell[l]{DivPrune \cite{divprune}}       & T & 0.46 & 1762 & 62.5 & 60.0 & 57.0 & 76.9 & \textbf{87.0} & 67.71 & 3.01 \\
        \makecell[l]{PDrop\textsuperscript{$\dagger$} \cite{pyramiddrop}} & T & 0.46 & \underline{1846} & 64.9 & 62.0 & 56.2 & 77.9 & 84.9 & 68.63 & 3.09 \\
        \makecell[l]{VisionZip\textsuperscript{$\dagger$} \cite{visionzip}}    & T & 0.46 & 1834 & 63.4 & 60.1 & \textbf{57.8} & 77.4 & 84.9 & 68.18 & 3.01 \\
        \makecell[l]{DART\textsuperscript{$\dagger$} \cite{dart}}              & T & 0.46 & 1829 & \textbf{66.3} & \underline{60.9} & 56.8 & \underline{78.2} & 85.3 & \underline{68.80} & 3.18 \\
        \makecell[l]{\textbf{CoMP\textsuperscript{$\dagger$} (Ours)}} & C & 0.46 & 1843 & \underline{66.1} & \textbf{61.9} & \underline{57.1} & \textbf{78.8} & \underline{85.7} & \textbf{69.23} & 2.94 \\
        \midrule
        \makecell[l]{FastV \cite{fastv}}             & T & 0.62 & 1256 & 48.0 & 46.1 & 47.8 & 55.0 & 48.0 & 48.29 & 2.38 \\
        \makecell[l]{SparseVLM \cite{sparsevlm}}     & T & 0.62 & 1589 & 60.1 & 53.8 & 53.4 & 68.2 & 77.5 & 61.63 & 2.15 \\
        \makecell[l]{PDrop \cite{pyramiddrop}}       & T & 0.62 & 1561 & 58.8 & 47.5 & 50.6 & 69.2 & 55.9 & 56.29 & 2.16 \\
        \makecell[l]{VisionZip \cite{visionzip}}     & T & 0.62 & 1690 & 60.1 & 55.1 & \underline{55.5} & 72.4 & 77.0 & 63.41 & 2.16 \\
        \makecell[l]{DART \cite{dart}}               & T & 0.62 & \underline{1765} & 60.6 & 55.9 & 54.4 & 72.4 & 73.9 & 63.37 & 2.38 \\
        \makecell[l]{DivPrune \cite{divprune}}       & T & 0.62 & 1674 & 59.3 & \underline{57.8} & 54.7 & 74.1 & \underline{85.6} & 65.20 & 2.16 \\
        \makecell[l]{PDrop\textsuperscript{$\dagger$} \cite{pyramiddrop}} & T & 0.62 & 1653 & 61.1 & 55.6 & 51.7 & 70.3 & 80.0 & 62.95 & 2.16 \\
        \makecell[l]{VisionZip\textsuperscript{$\dagger$} \cite{visionzip}} & T & 0.62 & 1756 & 61.5 & 57.0 & \textbf{56.0} & 74.2 & 80.9 & 65.39 & 2.16 \\
        \makecell[l]{DART\textsuperscript{$\dagger$} \cite{dart}} & T & 0.62 & \textbf{1823} & \textbf{64.7} & 57.1 & 54.7 & \underline{74.6} & 79.3 & \underline{65.92} & 2.38 \\
        \makecell[l]{\textbf{CoMP\textsuperscript{$\dagger$} (Ours)}} & C & 0.62 & 1735 & \underline{64.0} & \textbf{60.0} & 53.4 & \textbf{76.5} & \textbf{86.1} & \textbf{66.98} & 2.07 \\
        \bottomrule[1pt]
    \end{tabular}
\end{table*}

\paragraph{On Visual Question Answering Task}
We also evaluate on the visual question answering task using VQAv2 \cite{vqav2}.
Since the VQAv2 dataset does not provide a separate validation set \cite{vqav2}, we only utilize the CIM module in this setting. 
Despite this limitation, CoMP achieves a 0.3\% accuracy improvement over MADTP \cite{madtp}, demonstrating its generalizability.

\begin{table}
    \centering
    \caption{Ablation results of CIM and MPS on NLVR2 dataset at a pruning ratio of 0.8.}
    \label{tab:main_ablation_table}
    \footnotesize
    \begin{tabular}{c c| c  c| c}
    \toprule[1pt]
    CIM        & MPS        & Dev. Acc.      & Test Acc.      & GFLOPs \\
    \midrule
               &            & 76.69          & 77.88          & 26.39  \\
    \checkmark &            & 78.20          & 78.60          & 26.04  \\
    \checkmark & \checkmark & \textbf{79.23} & \textbf{79.62} & 25.97  \\
    \bottomrule[1pt]
    \end{tabular}
\end{table}

\begin{table}[!t]
\centering
\caption{(\textbf{Top}) Effect of CIP and CIT in CIM. CIP and CIT are token-weighted parameter importance and self-corrected token importance, respectively. (\textbf{Bottom}) Effect of CAS, RE and HI in MPS. CAS, RE and HI indicates pruning cost-aware mode shifting, random exploration and historical information, respectively. All experiments are conducted on NLVR2 at a pruning ratio of 0.8.}
\label{tab:ablation_all}
    \footnotesize
    \begin{tabular}{c | c@{\hskip 6pt}c@{\hskip 6pt}c| c@{\hskip 8pt}c@{\hskip 8pt}|c}
        \toprule[1pt]
        \multirow{5}{*}{CIM} & CIP & CIT & & Dev. Acc. & Test Acc. & GFLOPs \\
        \cline{2-7}
         & & & & 76.69 & 77.88 & 26.39 \\
         & \checkmark &  &  & 78.12 & 78.27 & 27.17 \\
         &  & \checkmark &  & 77.51 & 78.29 & 27.18 \\
         & \checkmark & \checkmark &  & \textbf{78.20} & \textbf{78.60} & 26.04 \\
         \cmidrule[1pt]{1-7}
        \multirow{4}{*}{MPS} & CAS & RE & HI & Dev. Acc. & Test Acc. & GFLOPs \\
        \cline{2-7}
         & \checkmark &  &  & 78.07 & 78.24 & 26.56 \\
         & \checkmark & \checkmark &  & 78.33 & 79.02 & 26.43 \\
         & \checkmark & \checkmark & \checkmark & \textbf{79.23} & \textbf{79.62} & 25.97 \\
        \bottomrule[1pt]
    \end{tabular}
\end{table}

\paragraph{Extension to LLM-based VLMs}
Our CoMP is originally designed for medium-scale VLMs fine-tuned on downstream tasks, where both validation and full-parameter training remain tolerable. To validate the effectiveness on recently prevailing LLM-based VLMs, we further evaluate CoMP on the representative LLaVA-v1.5-7B model.
We compare CoMP with the state-of-the-art vision token pruning methods \cite{fastv,sparsevlm,pyramiddrop,visionzip,dart,divprune}, except for CoreMatching \cite{corematching} due to incompletely released source code.
As demonstrated in~\cref{tab:llava_comparison}, by collaboratively pruning vision tokens, text tokens and parameters in the LLM component, CoMP surpasses the compared approaches on most benchmarks and achieves the highest average scores across all datasets. 
Notably, here CoMP completes compression within one epoch of supervised fine-tuning, which could help boost accuracy over training-free methods. For fairer comparisons, we also report the training-aware settings of PDrop \cite{pyramiddrop}, VisionZip \cite{visionzip} and DART \cite{dart}, denoted by `$\dagger$'. CoMP still improves the average scores of PDrop\textsuperscript{$\dagger$}, VisionZip\textsuperscript{$\dagger$} and DART\textsuperscript{$\dagger$} by 4.03\%, 1.59\% and 1.06\%, respectively, highlighting the contribution of our proposed multi-mode collaboration.

\subsection{Ablation Study}
\paragraph{On Main Components}
We evaluate the effect of the main components, including the Collaborative Importance Metric (CIM) and the Multi-Mode Pruning Strategy (MPS), on NLVR2 with BLIP model at a pruning ratio of 0.8. To that end, we establish a baseline by adopting simple joint pruning, where MADTP \cite{madtp} and UPop \cite{upop} are applied simultaneously without collaboration, as shown in~\cref{fig:summary}(iii).

As summarized in~\cref{tab:main_ablation_table}, CIM promotes Test Acc. by 0.72\% via mitigating the interference between the parameter and token importance metrics. The proposed MPS further improves the Test Acc. by 1.02\%, since it selects the optimal pruning mode during progressive pruning, based on current and historical pruning cost with random exploration.

\paragraph{On Effect of CIM}
We further evaluate the individual effect of the two sub-modules of CIM on NLVR2 at a pruning ratio
of 0.8. As shown in~\cref{tab:ablation_all} (Top), CIP that incorporates token importance into parameter importance metric and CIT that rectifies token importance by applying parameter pruning mask to attention weights improves Test Acc. by 0.39\% and 0.41\%, respectively.
These results imply that eliminating the interference between distinct pruning metrics achieves consistent performance gains.

\paragraph{On Effect of MPS}
We also evaluate the sub-modules of MPS on NLVR2 at a pruning ratio of 0.8.
As displayed in~\cref{tab:ablation_all} (Bottom), by applying CAS that directly replaces the original unified pruning process with five distinct modes and shifting with pure pruning cost leads to a slight performance drop. By additionally employing random exploration (RE) and historical information (HI), the test accuracy is improved by 0.78\% and 0.6\%, respectively. 
These results suggest that MPS accomplishes more stable pruning and avoids falling into local optimal mode by balancing short-term cost efficiency and random exploration.

\begin{figure}[!t]
    \centering
    \includegraphics[width=0.95\linewidth]{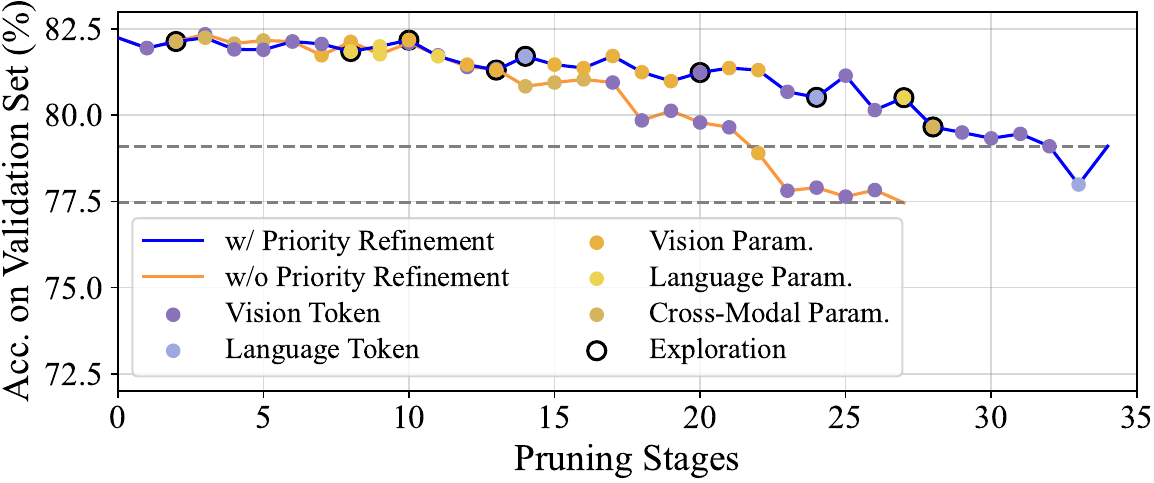}
    \caption{Visualization of pruning mode shifting in MPS with (w/) and without (w/o) priority refinement on NLVR2 at 0.8 pruning ratio. Colored points represent the next selected mode at each stage.}
    \label{fig:shift_stages}
    \vspace{-5pt}
\end{figure}

\paragraph{Visualization of Multi-Mode Shifting}
We further visualize the mode shifting and accuracy changes across all pruning stages. As shown in~\cref{fig:shift_stages}, by employing priority refinement with random exploration and historical information, our method exhibits more frequent mode shifting and explores diverse pruning modes at distinct stages. Even some necessary exploration steps contributes to recovery and ultimately leads to higher accuracy.
The curves also reveal a clear trend: token pruning dominates early stages, while parameter pruning becomes more prominent in later stages. This indicates that the redundancy initially lies in input data, whereas structural redundancy within models becomes more significant thereafter. This observation is consistent with the quantitative results in~\cref{tab:sota_table} and validates that the most prominent redundancy varies at different pruning stages.

\section{Conclusion}
\label{sec:conclusion}
In this paper, we investigate the joint pruning of parameters and tokens for VLMs, and propose a novel framework dubbed Collaborative Multi-Mode Pruning (CoMP).
We explicitly mitigate the inconsistency between the parameter and token importance by designing a Collaborative Importance Metric (CIM), and adaptively determine the optimal pruning mode during the progressive pruning progress via the Multi-Mode Pruning Strategy (MPS), thus enabling effective collaboration across distinct pruning modes.
Experimental results on diverse models and public datasets across multiple vision-language tasks demonstrate that our method consistently promotes the accuracy at high pruning ratios.
Our efforts highlight the potential of collaborative and mode-aware pruning schemes in achieving more effective and efficient compression of foundation vision-language models.

\clearpage
\section*{Acknowledgments}
This work was partly supported by the Beijing Natural Science Foundation (Nos. 4242044 and L259044), the CCF Baidu Open Fund, and the Fundamental Research Funds for the Central Universities.

{
    \small
    \bibliographystyle{ieeenat_fullname}
    \bibliography{main}

\begin{thebibliography}{68}
\providecommand{\natexlab}[1]{#1}
\providecommand{\url}[1]{\texttt{#1}}
\expandafter\ifx\csname urlstyle\endcsname\relax
  \providecommand{\doi}[1]{doi: #1}\else
  \providecommand{\doi}{doi: \begingroup \urlstyle{rm}\Url}\fi

\bibitem[Agrawal et~al.(2019)Agrawal, Desai, Wang, Chen, Jain, Johnson, Batra, Parikh, Lee, and Anderson]{nocaps}
Harsh Agrawal, Karan Desai, Yufei Wang, Xinlei Chen, Rishabh Jain, Mark Johnson, Dhruv Batra, Devi Parikh, Stefan Lee, and Peter Anderson.
\newblock Nocaps: Novel object captioning at scale.
\newblock In \emph{Proceedings of the IEEE/CVF International Conference on Computer Vision}, pages 8947--8956, 2019.

\bibitem[Alvar et~al.(2025)Alvar, Singh, Akbari, and Zhang]{divprune}
Saeed~Ranjbar Alvar, Gursimran Singh, Mohammad Akbari, and Yong Zhang.
\newblock Divprune: Diversity-based visual token pruning for large multimodal models.
\newblock In \emph{Proceedings of the IEEE/CVF Conference on Computer Vision and Pattern Recognition}, pages 9392--9401, 2025.

\bibitem[An et~al.(2024)An, Zhao, Yu, Tang, and Wang]{flap}
Yongqi An, Xu Zhao, Tao Yu, Ming Tang, and Jinqiao Wang.
\newblock Fluctuation-based adaptive structured pruning for large language models.
\newblock In \emph{Proceedings of the AAAI Conference on Artificial Intelligence}, number~10, pages 10865--10873, 2024.

\bibitem[Anderson et~al.(2016)Anderson, Fernando, Johnson, and Gould]{spice}
Peter Anderson, Basura Fernando, Mark Johnson, and Stephen Gould.
\newblock Spice: Semantic propositional image caption evaluation.
\newblock In \emph{Proceedings of the European Conference on Computer Vision}, pages 382--398, 2016.

\bibitem[Anwar et~al.(2017)Anwar, Hwang, and Sung]{structured_prune_survey}
Sajid Anwar, Kyuyeon Hwang, and Wonyong Sung.
\newblock Structured pruning of deep convolutional neural networks.
\newblock \emph{ACM Journal on Emerging Technologies in Computing Systems}, 13\penalty0 (3):\penalty0 1--18, 2017.

\bibitem[Awais et~al.(2025)Awais, Naseer, Khan, Anwer, Cholakkal, Shah, Yang, and Khan]{foundation_model_survey}
Muhammad Awais, Muzammal Naseer, Salman Khan, Rao~Muhammad Anwer, Hisham Cholakkal, Mubarak Shah, Ming-Hsuan Yang, and Fahad~Shahbaz Khan.
\newblock Foundation models defining a new era in vision: a survey and outlook.
\newblock \emph{IEEE Transactions on Pattern Analysis and Machine Intelligence}, 47\penalty0 (4):\penalty0 2245--2264, 2025.

\bibitem[Bian et~al.(2023)Bian, Wang, Han, and Wang]{mstm}
Zhe Bian, Zhe Wang, Wenqiang Han, and Kangping Wang.
\newblock Multi-scale and token mergence: Make your vit more efficient.
\newblock \emph{arXiv preprint arXiv:2306.04897}, 2023.

\bibitem[Bolya et~al.(2023)Bolya, Fu, Dai, Zhang, Feichtenhofer, and Hoffman]{tome}
Daniel Bolya, Cheng-Yang Fu, Xiaoliang Dai, Peizhao Zhang, Christoph Feichtenhofer, and Judy Hoffman.
\newblock Token merging: Your vit but faster.
\newblock In \emph{Proceedings of the International Conference on Learning Representations}, 2023.

\bibitem[Cao et~al.(2024)Cao, Ye, Li, Yu, Tang, Lu, and Chen]{madtp}
Jianjian Cao, Peng Ye, Shengze Li, Chong Yu, Yansong Tang, Jiwen Lu, and Tao Chen.
\newblock Madtp: Multimodal alignment-guided dynamic token pruning for accelerating vision-language transformer.
\newblock In \emph{Proceedings of the IEEE/CVF Conference on Computer Vision and Pattern Recognition}, pages 15710--15719, 2024.

\bibitem[Chavan et~al.(2022)Chavan, Shen, Liu, Liu, Cheng, and Xing]{vit_slim}
Arnav Chavan, Zhiqiang Shen, Zhuang Liu, Zechun Liu, Kwang-Ting Cheng, and Eric~P Xing.
\newblock Vision transformer slimming: Multi-dimension searching in continuous optimization space.
\newblock In \emph{Proceedings of the IEEE/CVF Conference on Computer Vision and Pattern Recognition}, pages 4931--4941, 2022.

\bibitem[Chen et~al.(2024{\natexlab{a}})Chen, Zhao, Liu, Bai, Lin, Zhou, and Chang]{fastv}
Liang Chen, Haozhe Zhao, Tianyu Liu, Shuai Bai, Junyang Lin, Chang Zhou, and Baobao Chang.
\newblock An image is worth 1/2 tokens after layer 2: Plug-and-play inference acceleration for large vision-language models.
\newblock In \emph{Proceedings of the European Conference on Computer Vision}, pages 19--35, 2024{\natexlab{a}}.

\bibitem[Chen et~al.(2024{\natexlab{b}})Chen, Fang, Ma, and Wang]{slimsam}
Zigeng Chen, Gongfan Fang, Xinyin Ma, and Xinchao Wang.
\newblock Slimsam: 0.1\% data makes segment anything slim.
\newblock In \emph{Proceedings of the Annual Conference on Neural Information Processing Systems}, pages 39434--39461, 2024{\natexlab{b}}.

\bibitem[Cheng et~al.(2024)Cheng, Zhang, and Shi]{prune_survey}
Hongrong Cheng, Miao Zhang, and Javen~Qinfeng Shi.
\newblock A survey on deep neural network pruning: Taxonomy, comparison, analysis, and recommendations.
\newblock \emph{IEEE Transactions on Pattern Analysis and Machine Intelligence}, 46\penalty0 (12):\penalty0 10558--10578, 2024.

\bibitem[Cubuk et~al.(2020)Cubuk, Zoph, Shlens, and Le]{randomaug}
Ekin~D Cubuk, Barret Zoph, Jonathon Shlens, and Quoc~V Le.
\newblock Randaugment: Practical automated data augmentation with a reduced search space.
\newblock In \emph{Proceedings of the IEEE/CVF Conference on Computer Vision and Pattern Recognition}, pages 702--703, 2020.

\bibitem[Fang et~al.(2024)Fang, Ma, Mi, and Wang]{isomorphic}
Gongfan Fang, Xinyin Ma, Michael~Bi Mi, and Xinchao Wang.
\newblock Isomorphic pruning for vision models.
\newblock In \emph{Proceedings of the European Conference on Computer Vision}, pages 232--250, 2024.

\bibitem[Fu et~al.(2023)Fu, Chen, Shen, Qin, Zhang, Lin, Yang, Zheng, Li, Sun, et~al.]{mme}
Chaoyou Fu, Peixian Chen, Yunhang Shen, Yulei Qin, Mengdan Zhang, Xu Lin, Jinrui Yang, Xiawu Zheng, Ke Li, Xing Sun, et~al.
\newblock Mme: A comprehensive evaluation benchmark for multimodal large language models.
\newblock In \emph{Proceedings of the Annual Conference on Neural Information Processing Systems Datasets and Benchmarks Track}, 2023.

\bibitem[Goyal et~al.(2020)Goyal, Choudhury, Raje, Chakaravarthy, Sabharwal, and Verma]{power_bert}
Saurabh Goyal, Anamitra~Roy Choudhury, Saurabh Raje, Venkatesan Chakaravarthy, Yogish Sabharwal, and Ashish Verma.
\newblock Power-bert: Accelerating bert inference via progressive word-vector elimination.
\newblock In \emph{Proceedings of the International Conference on Machine Learning}, pages 3690--3699, 2020.

\bibitem[Goyal et~al.(2017)Goyal, Khot, Summers-Stay, Batra, and Parikh]{vqav2}
Yash Goyal, Tejas Khot, Douglas Summers-Stay, Dhruv Batra, and Devi Parikh.
\newblock Making the v in vqa matter: Elevating the role of image understanding in visual question answering.
\newblock In \emph{Proceedings of the IEEE/CVF Conference on Computer Vision and Pattern Recognition}, pages 6904--6913, 2017.

\bibitem[Guo et~al.(2023)Guo, Zhang, Wong, Nie, and Kankanhalli]{elip}
Yangyang Guo, Haoyu Zhang, Yongkang Wong, Liqiang Nie, and Mohan Kankanhalli.
\newblock Elip: Efficient language-image pre-training with fewer vision tokens.
\newblock \emph{arXiv preprint arXiv:2309.16738}, 2023.

\bibitem[Hudson and Manning(2019)]{gqa}
Drew~A Hudson and Christopher~D Manning.
\newblock Gqa: A new dataset for real-world visual reasoning and compositional question answering.
\newblock In \emph{Proceedings of the IEEE/CVF Conference on Computer Vision and Pattern Recognition}, pages 6700--6709, 2019.

\bibitem[Jiang et~al.(2023)Jiang, Chen, Huang, and Wang]{miep}
Liangwei Jiang, Jiaxin Chen, Di Huang, and Yunhong Wang.
\newblock Miep: Channel pruning with multi-granular importance estimation for object detection.
\newblock In \emph{Proceedings of the ACM International Conference on Multimedia}, pages 2908--2917, 2023.

\bibitem[Ju et~al.(2024)Ju, Wang, Cheng, Chen, Zhai, Huang, Lan, Xiao, and Zheng]{turbo}
Chen Ju, Haicheng Wang, Haozhe Cheng, Xu Chen, Zhonghua Zhai, Weilin Huang, Jinsong Lan, Shuai Xiao, and Bo Zheng.
\newblock Turbo: Informativity-driven acceleration plug-in for vision-language large models.
\newblock In \emph{Proceedings of the European Conference on Computer Vision}, pages 436--455, 2024.

\bibitem[Kong et~al.(2022)Kong, Dong, Ma, Meng, Niu, Sun, Shen, Yuan, Ren, Tang, et~al.]{spvit}
Zhenglun Kong, Peiyan Dong, Xiaolong Ma, Xin Meng, Wei Niu, Mengshu Sun, Xuan Shen, Geng Yuan, Bin Ren, Hao Tang, et~al.
\newblock Spvit: Enabling faster vision transformers via latency-aware soft token pruning.
\newblock In \emph{Proceedings of the European Conference on Computer Vision}, pages 620--640, 2022.

\bibitem[Li et~al.(2022)Li, Li, Xiong, and Hoi]{blip}
Junnan Li, Dongxu Li, Caiming Xiong, and Steven Hoi.
\newblock Blip: Bootstrapping language-image pre-training for unified vision-language understanding and generation.
\newblock In \emph{Proceedings of the International Conference on Machine Learning}, pages 12888--12900, 2022.

\bibitem[Li et~al.(2023{\natexlab{a}})Li, Zhang, Xu, Wang, Yan, Xia, Yang, Cao, Sun, Deng, et~al.]{constraint_token_prune}
Junyan Li, Li~Lyna Zhang, Jiahang Xu, Yujing Wang, Shaoguang Yan, Yunqing Xia, Yuqing Yang, Ting Cao, Hao Sun, Weiwei Deng, et~al.
\newblock Constraint-aware and ranking-distilled token pruning for efficient transformer inference.
\newblock In \emph{Proceedings of the ACM SIGKDD Conference on Knowledge Discovery and Data Mining}, pages 1280--1290, 2023{\natexlab{a}}.

\bibitem[Li et~al.(2025)Li, Hu, Ning, Liu, Hong, Jia, Li, Yan, Ran, Dai, et~al.]{mbq}
Shiyao Li, Yingchun Hu, Xuefei Ning, Xihui Liu, Ke Hong, Xiaotao Jia, Xiuhong Li, Yaqi Yan, Pei Ran, Guohao Dai, et~al.
\newblock Mbq: Modality-balanced quantization for large vision-language models.
\newblock In \emph{Proceedings of the IEEE/CVF Conference on Computer Vision and Pattern Recognition}, pages 4167--4177, 2025.

\bibitem[Li et~al.(2023{\natexlab{b}})Li, Du, Zhou, Wang, Zhao, and Wen]{pope}
Yifan Li, Yifan Du, Kun Zhou, Jinpeng Wang, Wayne~Xin Zhao, and Ji-Rong Wen.
\newblock Evaluating object hallucination in large vision-language models.
\newblock In \emph{Proceedings of the Conference on Empirical Methods in Natural Language Processing}, pages 292--305, 2023{\natexlab{b}}.

\bibitem[Li et~al.(2026)Li, Tang, Meng, Fan, Chai, Ma, Wang, and Zhu]{prance}
Ye Li, Chen Tang, Yuan Meng, Jiajun Fan, Zenghao Chai, Xinzhu Ma, Zhi Wang, and Wenwu Zhu.
\newblock Prance: Joint token-optimization and structural channel-pruning for adaptive vit inference.
\newblock \emph{IEEE Transactions on Pattern Analysis and Machine Intelligence}, 48\penalty0 (1):\penalty0 283--298, 2026.

\bibitem[Liang et~al.(2021)Liang, Glossner, Wang, Shi, and Zhang]{prune_quantize_survey}
Tailin Liang, John Glossner, Lei Wang, Shaobo Shi, and Xiaotong Zhang.
\newblock Pruning and quantization for deep neural network acceleration: A survey.
\newblock \emph{Neurocomputing}, 461:\penalty0 370--403, 2021.

\bibitem[Liang et~al.(2022)Liang, Chongjian, Tong, Song, Wang, and Xie]{evit}
Youwei Liang, GE Chongjian, Zhan Tong, Yibing Song, Jue Wang, and Pengtao Xie.
\newblock Evit: Expediting vision transformers via token reorganizations.
\newblock In \emph{Proceedings of the International Conference on Learning Representations}, 2022.

\bibitem[Lin et~al.(2024)Lin, Bai, Liu, Hou, Sun, Song, Wei, and Sun]{mope}
Haokun Lin, Haoli Bai, Zhili Liu, Lu Hou, Muyi Sun, Linqi Song, Ying Wei, and Zhenan Sun.
\newblock Mope-clip: Structured pruning for efficient vision-language models with module-wise pruning error metric.
\newblock In \emph{Proceedings of the IEEE/CVF Conference on Computer Vision and Pattern Recognition}, pages 27370--27380, 2024.

\bibitem[Lin et~al.(2014)Lin, Maire, Belongie, Hays, Perona, Ramanan, Doll{\'a}r, and Zitnick]{coco}
Tsung-Yi Lin, Michael Maire, Serge Belongie, James Hays, Pietro Perona, Deva Ramanan, Piotr Doll{\'a}r, and C~Lawrence Zitnick.
\newblock Microsoft coco: Common objects in context.
\newblock In \emph{Proceedings of the European Conference on Computer Vision}, pages 740--755, 2014.

\bibitem[Liu et~al.(2023)Liu, Li, Wu, and Lee]{llava}
Haotian Liu, Chunyuan Li, Qingyang Wu, and Yong~Jae Lee.
\newblock Visual instruction tuning.
\newblock In \emph{Proceedings of the Annual Conference on Neural Information Processing Systems}, pages 34892--34916, 2023.

\bibitem[Liu et~al.(2024{\natexlab{a}})Liu, Li, Li, and Lee]{llavaimproved}
Haotian Liu, Chunyuan Li, Yuheng Li, and Yong~Jae Lee.
\newblock Improved baselines with visual instruction tuning.
\newblock In \emph{Proceedings of the IEEE/CVF Conference on Computer Vision and Pattern Recognition}, pages 26296--26306, 2024{\natexlab{a}}.

\bibitem[Liu et~al.(2021)Liu, Zhang, Kuang, Zhou, Xue, Wang, Chen, Yang, Liao, and Zhang]{group_fisher}
Liyang Liu, Shilong Zhang, Zhanghui Kuang, Aojun Zhou, Jing-Hao Xue, Xinjiang Wang, Yimin Chen, Wenming Yang, Qingmin Liao, and Wayne Zhang.
\newblock Group fisher pruning for practical network compression.
\newblock In \emph{Proceedings of the International Conference on Machine Learning}, pages 7021--7032, 2021.

\bibitem[Liu et~al.(2024{\natexlab{b}})Liu, Duan, Zhang, Li, Zhang, Zhao, Yuan, Wang, He, Liu, et~al.]{mmbench}
Yuan Liu, Haodong Duan, Yuanhan Zhang, Bo Li, Songyang Zhang, Wangbo Zhao, Yike Yuan, Jiaqi Wang, Conghui He, Ziwei Liu, et~al.
\newblock Mmbench: Is your multi-modal model an all-around player?
\newblock In \emph{Proceedings of the European Conference on Computer Vision}, pages 216--233, 2024{\natexlab{b}}.

\bibitem[Loshchilov and Hutter(2016)]{cos_lr}
Ilya Loshchilov and Frank Hutter.
\newblock Sgdr: Stochastic gradient descent with warm restarts.
\newblock \emph{arXiv preprint arXiv:1608.03983}, 2016.

\bibitem[Loshchilov and Hutter(2017)]{adamw}
Ilya Loshchilov and Frank Hutter.
\newblock Decoupled weight decay regularization.
\newblock \emph{arXiv preprint arXiv:1711.05101}, 2017.

\bibitem[Ma et~al.(2023)Ma, Fang, and Wang]{llmpruner}
Xinyin Ma, Gongfan Fang, and Xinchao Wang.
\newblock Llm-pruner: On the structural pruning of large language models.
\newblock In \emph{Proceedings of the Annual Conference on Neural Information Processing Systems}, pages 21702--21720, 2023.

\bibitem[Radford et~al.(2021)Radford, Kim, Hallacy, Ramesh, Goh, Agarwal, Sastry, Askell, Mishkin, Clark, et~al.]{clip}
Alec Radford, Jong~Wook Kim, Chris Hallacy, Aditya Ramesh, Gabriel Goh, Sandhini Agarwal, Girish Sastry, Amanda Askell, Pamela Mishkin, Jack Clark, et~al.
\newblock Learning transferable visual models from natural language supervision.
\newblock In \emph{Proceedings of the International Conference on Machine Learning}, pages 8748--8763, 2021.

\bibitem[Rao et~al.(2021)Rao, Zhao, Liu, Lu, Zhou, and Hsieh]{dynamicvit}
Yongming Rao, Wenliang Zhao, Benlin Liu, Jiwen Lu, Jie Zhou, and Cho-Jui Hsieh.
\newblock Dynamicvit: Efficient vision transformers with dynamic token sparsification.
\newblock In \emph{Proceedings of the Annual Conference on Neural Information Processing Systems}, pages 13937--13949, 2021.

\bibitem[Shi et~al.(2023)Shi, Tao, Jin, Yang, Yuan, and Wang]{upop}
Dachuan Shi, Chaofan Tao, Ying Jin, Zhendong Yang, Chun Yuan, and Jiaqi Wang.
\newblock Upop: Unified and progressive pruning for compressing vision-language transformers.
\newblock In \emph{Proceedings of the International Conference on Machine Learning}, pages 31292--31311, 2023.

\bibitem[Shi et~al.(2024)Shi, Tao, Rao, Yang, Yuan, and Wang]{crossget}
Dachuan Shi, Chaofan Tao, Anyi Rao, Zhendong Yang, Chun Yuan, and Jiaqi Wang.
\newblock Crossget: Cross-guided ensemble of tokens for accelerating vision-language transformers.
\newblock In \emph{Proceedings of the International Conference on Machine Learning}, pages 44960--44990, 2024.

\bibitem[Singh et~al.(2019)Singh, Natarajan, Shah, Jiang, Chen, Batra, Parikh, and Rohrbach]{textvqa}
Amanpreet Singh, Vivek Natarajan, Meet Shah, Yu Jiang, Xinlei Chen, Dhruv Batra, Devi Parikh, and Marcus Rohrbach.
\newblock Towards vqa models that can read.
\newblock In \emph{Proceedings of the IEEE/CVF Conference on Computer Vision and Pattern Recognition}, pages 8317--8326, 2019.

\bibitem[Suhr et~al.(2019)Suhr, Zhou, Zhang, Zhang, Bai, and Artzi]{nlvr2}
Alane Suhr, Stephanie Zhou, Ally Zhang, Iris Zhang, Huajun Bai, and Yoav Artzi.
\newblock A corpus for reasoning about natural language grounded in photographs.
\newblock In \emph{Proceedings of the Annual Meeting of the Association for Computational Linguistics}, pages 6418--6428, 2019.

\bibitem[Sun et~al.(2023)Sun, Liu, Bair, and Kolter]{wanda}
Mingjie Sun, Zhuang Liu, Anna Bair, and J~Zico Kolter.
\newblock A simple and effective pruning approach for large language models.
\newblock In \emph{Proceedings of the International Conference on Learning Representations}, 2023.

\bibitem[Tang et~al.(2024)Tang, Wang, Guo, Tu, Han, Hu, and Tao]{transformer_compression_survey}
Yehui Tang, Yunhe Wang, Jianyuan Guo, Zhijun Tu, Kai Han, Hailin Hu, and Dacheng Tao.
\newblock A survey on transformer compression.
\newblock \emph{arXiv preprint arXiv:2402.05964}, 2024.

\bibitem[Vaswani et~al.(2017)Vaswani, Shazeer, Parmar, Uszkoreit, Jones, Gomez, Kaiser, and Polosukhin]{transformer}
Ashish Vaswani, Noam Shazeer, Niki Parmar, Jakob Uszkoreit, Llion Jones, Aidan~N Gomez, {\L}ukasz Kaiser, and Illia Polosukhin.
\newblock Attention is all you need.
\newblock In \emph{Proceedings of the Annual Conference on Neural Information Processing Systems}, pages 5998--6008, 2017.

\bibitem[Vedantam et~al.(2015)Vedantam, Lawrence~Zitnick, and Parikh]{cider}
Ramakrishna Vedantam, C Lawrence~Zitnick, and Devi Parikh.
\newblock Cider: Consensus-based image description evaluation.
\newblock In \emph{Proceedings of the IEEE/CVF Conference on Computer Vision and Pattern Recognition}, pages 4566--4575, 2015.

\bibitem[Wang et~al.(2021)Wang, Zhang, and Han]{spatten}
Hanrui Wang, Zhekai Zhang, and Song Han.
\newblock Spatten: Efficient sparse attention architecture with cascade token and head pruning.
\newblock In \emph{Proceedings of the IEEE International Symposium on High-Performance Computer Architecture}, pages 97--110, 2021.

\bibitem[Wang et~al.(2024)Wang, Dedhia, and Jha]{zero_tprune}
Hongjie Wang, Bhishma Dedhia, and Niraj~K Jha.
\newblock Zero-tprune: Zero-shot token pruning through leveraging of the attention graph in pre-trained transformers.
\newblock In \emph{Proceedings of the IEEE/CVF Conference on Computer Vision and Pattern Recognition}, pages 16070--16079, 2024.

\bibitem[Wang et~al.(2025)Wang, Ye, Chung, Liu, Lin, Kuo, Ma, Zhang, and Chen]{corematching}
Qinsi Wang, Hancheng Ye, Ming-Yu Chung, Yudong Liu, Yueqian Lin, Martin Kuo, Mingyuan Ma, Jianyi Zhang, and Yiran Chen.
\newblock Corematching: A co-adaptive sparse inference framework with token and neuron pruning for comprehensive acceleration of vision-language models.
\newblock In \emph{Proceedings of the International Conference on Machine Learning}, 2025.

\bibitem[Wei et~al.(2023)Wei, Ye, Zhang, Tang, and Liang]{joint_token_prune_squeeze}
Siyuan Wei, Tianzhu Ye, Shen Zhang, Yao Tang, and Jiajun Liang.
\newblock Joint token pruning and squeezing towards more aggressive compression of vision transformers.
\newblock In \emph{Proceedings of the IEEE/CVF Conference on Computer Vision and Pattern Recognition}, pages 2092--2101, 2023.

\bibitem[Wen et~al.(2025)Wen, Gao, Wang, Zhang, Zhang, Li, He, and Zhang]{dart}
Zichen Wen, Yifeng Gao, Shaobo Wang, Junyuan Zhang, Qintong Zhang, Weijia Li, Conghui He, and Linfeng Zhang.
\newblock Stop looking for “important tokens” in multimodal language models: Duplication matters more.
\newblock In \emph{Proceedings of the Conference on Empirical Methods in Natural Language Processing}, pages 9972--9991, 2025.

\bibitem[Wu et~al.(2023{\natexlab{a}})Wu, Peng, Zhou, Xiao, Liu, Yuan, Xuan, Valenzuela, Chen, Wang, et~al.]{tinyclip}
Kan Wu, Houwen Peng, Zhenghong Zhou, Bin Xiao, Mengchen Liu, Lu Yuan, Hong Xuan, Michael Valenzuela, Xi~Stephen Chen, Xinggang Wang, et~al.
\newblock Tinyclip: Clip distillation via affinity mimicking and weight inheritance.
\newblock In \emph{Proceedings of the IEEE/CVF International Conference on Computer Vision}, pages 21970--21980, 2023{\natexlab{a}}.

\bibitem[Wu et~al.(2023{\natexlab{b}})Wu, Zeng, Wang, and Chen]{ppt}
Xinjian Wu, Fanhu Zeng, Xiudong Wang, and Xinghao Chen.
\newblock Ppt: Token pruning and pooling for efficient vision transformers.
\newblock \emph{arXiv preprint arXiv:2310.01812}, 2023{\natexlab{b}}.

\bibitem[Wu et~al.(2023{\natexlab{c}})Wu, Chen, and Wang]{samp}
Zimeng Wu, Jiaxin Chen, and Yunhong Wang.
\newblock Samp: Sub-task aware model pruning with layer-wise channel balancing for person search.
\newblock In \emph{Chinese Conference on Pattern Recognition and Computer Vision}, pages 199--211, 2023{\natexlab{c}}.

\bibitem[Wu et~al.(2025)Wu, Chen, and Wang]{ukmp}
Zimeng Wu, Jiaxin Chen, and Yunhong Wang.
\newblock Unified knowledge maintenance pruning and progressive recovery with weight recalling for large vision-language models.
\newblock In \emph{Proceedings of the AAAI Conference on Artificial Intelligence}, pages 8550--8558, 2025.

\bibitem[Wu et~al.(2026)Wu, Wang, Jin, Chen, and Wang]{spts}
Zimeng Wu, Donghao Wang, Chaozhe Jin, Jiaxin Chen, and Yunhong Wang.
\newblock Probe and skip: Self-predictive token skipping for efficient long-context llm inference.
\newblock \emph{arXiv preprint arXiv:2601.13155}, 2026.

\bibitem[Xing et~al.(2025)Xing, Huang, Dong, Lu, Zhang, Zang, Cao, He, Wang, Wu, et~al.]{pyramiddrop}
Long Xing, Qidong Huang, Xiaoyi Dong, Jiajie Lu, Pan Zhang, Yuhang Zang, Yuhang Cao, Conghui He, Jiaqi Wang, Feng Wu, et~al.
\newblock Conical visual concentration for efficient large vision-language models.
\newblock In \emph{Proceedings of the IEEE/CVF Conference on Computer Vision and Pattern Recognition}, pages 14593--14603, 2025.

\bibitem[Xu et~al.(2016)Xu, Mei, Yao, and Rui]{msrvtt}
Jun Xu, Tao Mei, Ting Yao, and Yong Rui.
\newblock Msr-vtt: A large video description dataset for bridging video and language.
\newblock In \emph{Proceedings of the IEEE/CVF Conference on Computer Vision and Pattern Recognition}, pages 5288--5296, 2016.

\bibitem[Xu et~al.(2024)Xu, Yin, Cai, Yi, Xu, Wang, Wu, Zhao, Yang, Wang, et~al.]{multimodal_foundation_survey}
Mengwei Xu, Wangsong Yin, Dongqi Cai, Rongjie Yi, Daliang Xu, Qipeng Wang, Bingyang Wu, Yihao Zhao, Chen Yang, Shihe Wang, et~al.
\newblock A survey of resource-efficient llm and multimodal foundation models.
\newblock \emph{arXiv preprint arXiv:2401.08092}, 2024.

\bibitem[Yang et~al.(2025)Yang, Chen, Tian, Wang, Li, Yu, and Jia]{visionzip}
Senqiao Yang, Yukang Chen, Zhuotao Tian, Chengyao Wang, Jingyao Li, Bei Yu, and Jiaya Jia.
\newblock Visionzip: Longer is better but not necessary in vision language models.
\newblock In \emph{Proceedings of the IEEE/CVF Conference on Computer Vision and Pattern Recognition}, pages 19792--19802, 2025.

\bibitem[Yin et~al.(2022)Yin, Vahdat, Alvarez, Mallya, Kautz, and Molchanov]{a_vit}
Hongxu Yin, Arash Vahdat, Jose~M Alvarez, Arun Mallya, Jan Kautz, and Pavlo Molchanov.
\newblock A-vit: Adaptive tokens for efficient vision transformer.
\newblock In \emph{Proceedings of the IEEE/CVF Conference on Computer Vision and Pattern Recognition}, pages 10809--10818, 2022.

\bibitem[Young et~al.(2014)Young, Lai, Hodosh, and Hockenmaier]{flickr30k}
Peter Young, Alice Lai, Micah Hodosh, and Julia Hockenmaier.
\newblock From image descriptions to visual denotations: New similarity metrics for semantic inference over event descriptions.
\newblock \emph{Transactions of the Association for Computational Linguistics}, 2:\penalty0 67--78, 2014.

\bibitem[Zhang et~al.(2026{\natexlab{a}})Zhang, Chen, Guo, and Di]{jiayi}
Jiayi Zhang, Jiaxin Chen, Xiefan Guo, and Huang Di.
\newblock Towards accurate quantization for large vision-language models via zeroth-order gradient optimization and sectioned logarithmic quantizer.
\newblock In \emph{Proceedings of the International Conference on Acoustics, Speech, and Signal Processing}, 2026{\natexlab{a}}.

\bibitem[Zhang et~al.(2026{\natexlab{b}})Zhang, Wang, Wang, and Chen]{mingfang}
Mingfang Zhang, Yunhong Wang, Lu Wang, and Jiaxin Chen.
\newblock Parameter-efficient adaptation for mllms via implicit modality decomposition.
\newblock In \emph{Proceedings of the IEEE/CVF Conference on Computer Vision and Pattern Recognition}, 2026{\natexlab{b}}.

\bibitem[Zhang et~al.(2025)Zhang, Fan, Ma, Zheng, Huang, Cheng, Gudovskiy, Okuno, Nakata, Keutzer, et~al.]{sparsevlm}
Yuan Zhang, Chun-Kai Fan, Junpeng Ma, Wenzhao Zheng, Tao Huang, Kuan Cheng, Denis Gudovskiy, Tomoyuki Okuno, Yohei Nakata, Kurt Keutzer, et~al.
\newblock Sparsevlm: Visual token sparsification for efficient vision-language model inference.
\newblock In \emph{Proceedings of the International Conference on Machine Learning}, 2025.

\end{thebibliography}
}

 \clearpage
\setcounter{page}{1}
\setcounter{section}{0}
\renewcommand\thesection{\Alph{section}}
\setcounter{figure}{0}
\setcounter{table}{0}
\renewcommand{\thefigure}{\Alph{figure}}
\renewcommand{\thetable}{\Alph{table}}
\maketitlesupplementary

In this document, we additionally provide the overall algorithm of our CoMP method in~\cref{sec:pseudocode}, detailed descriptions of the datasets and evaluation metrics in~\cref{sec:desc_dataset_metric}, more implementation details in~\cref{sec:detailed_implementation}, detailed description of dimension mapping for parameter pruning in~\cref{sec:dim_map}, and additional experimental results in~\cref{sec:more_experiments}. 
Specifically, we evaluate real-world inference latency in~\cref{sec:inference_latency}, demonstrate contributions of difference modes in~\cref{sec:mode_contribution}, include more ablations on hyperparameters in~\cref{sec:hyperparameter_ablation}, multi-seed statistical results in~\cref{sec:more_seeds}, video-text performance in~\cref{sec:evaluation_video}, further validation of orthogonality to single-mode pruning in~\cref{sec:orthogonality_param} and discussion on computational overhead in~\cref{sec:computational_overhead}. Besides, we provide supplementary observations on the inconsistency between parameter and token importance in~~\cref{sec:more_observation}.

\section{Overall Algorithm of CoMP}
\label{sec:pseudocode}

\begin{algorithm*}[t]
    \small
    \caption{\small Collaborative Multi-Mode Pruning (CoMP)}
    \label{psudocode}
    \setcounter{AlgoLine}{0}
    \LinesNumbered

    \KwIn{Uncompressed VLM $\mathcal{F}(\cdot|\bm{W})$ in full parameters $\bm{W}$ with $L$ layers, loss function $\mathcal{L}$, target FLOPs $\mathit{TF}$, pruning modes $\mathcal{B}$, probability for exploration $\rho$, interval steps $\mathcal{I}$ between mode shifting, training dataset $\mathcal{D}_{\text{train}}$, and validation dataset $\mathcal{D}_{\text{val}}$.}

    \KwOut{Pruned model $\mathcal{F}(\cdot|\hat{\bm{W}}, \hat{\theta_v^t}, \hat{\theta_l^t})$ with parameters $\hat{\bm{W}}$ and thresholds $\hat{\theta_v^t}, \hat{\theta_l^t}$ for token pruning.}

    \BlankLine
    \textcolor{gray}{\texttt{\# Initialize model with parameter masks $\bm{M}^p$ and token thresholds $\theta_v^t, \theta_l^t$ for pruning}} \\
    $\mathcal{F}\gets \mathcal{F}(\cdot|\bm{W}, \bm{M}^p, \theta_v^t, \theta_l^t)$
    
    $\bm{M}^p\gets \mathbf{1}$~~
    \textcolor{gray}{\texttt{\# Initialize parameter pruning masks to 0}} 

    $\mathcal{J}\gets\mathcal{I}/6$~~
    \textcolor{gray}{\texttt{\# Interval between parameter pruning mask updates}} 
    
    \textcolor{gray}{\texttt{\# Initialize the MPS module.~Elements of $\Theta$ correspond to $\theta^p_v, \theta^p_l, \theta^p_c, \theta^t_v, \theta^t_l$ in order}} \\
    $\Theta \gets \{0, 0, 0, 0, 0\}$, $\mathcal{R} \gets \{0, 0, 0, 0, 0\}$, $\mathcal{T}\gets \{0,0,0,0,0\}$, $m\gets 0$, $T\gets 0$ 
    
    $\mathit{CA},~\mathit{CF}\gets \text{Evaluate}(\mathcal{F}, \mathcal{D}_{\text{val}})$~~
    \textcolor{gray}{\texttt{\# Initialize accuracy $\mathit{CA}$ and FLOPs $\mathit{CF}$}}
    
    \While{$\mathit{CF} > \mathit{TF}$}{
        $\bm{S}'^p\gets 0$~~
        \textcolor{gray}{\texttt{\# Reset the accumulated parameter importance}}
        
        \textcolor{gray}{\texttt{\# Conduct the current pruning mode $\mathcal{B}_m$ by adjusting the threshold}} \\
        $\Theta_m \gets \Theta_m + \Delta\Theta_m$, $T\gets T+1$

        \textcolor{gray}{\texttt{\# Optimize the model for $\mathcal{I}$ steps at current stage}}\\
        \For{$i \gets 1$ \KwTo $\mathcal{I}$}{
            $(\bm{X}^0,Y)\gets \text{Sample}(\mathcal{D}_{\text{train}})$~~
            \textcolor{gray}{\texttt{\# Sample data from the training set}}
            
            \For{$l\gets 1$ \KwTo $L$}{ 
                \textcolor{gray}{\texttt{\# Run layer-wise forward propagation with partially masked parameters}} \\
                $\bm{X}^l \gets \mathcal{F}^l(\bm{X}^{l-1}|\bm{W}^l, \bm{M}^p)$

                \textcolor{gray}{\texttt{\# Calculate the self-corrected token importance by CIM}} \\
                $\bm{S}^{t, l} \gets \text{CIT}(\bm{X}^l)$

                \textcolor{gray}{\texttt{\# Perform layer-wise token pruning with $\theta^t\in\{\theta^t_v, \theta^t_l\}$}} \\
                $\bm{M}^{t,l}\gets \mathbb{I}(\bm{S}^{t,l}> \theta^t)$,~$\bm{X}^l \gets \bm{X}^l\odot \bm{M}^{t,l}$
            }

            \If{$\mathcal{B}_m\in \{B^p_v, B^p_l, B^p_c\}$}{
                \textcolor{gray}{\texttt{\# Follow the pruning framework in \cite{upop}: accumulate importance every $\mathcal{J}$ steps and uniformly decay the mask to $0$ for stability.}}\\
               $\bm{S}'^p \gets \bm{S}'^p + \text{CIP}(\bm{S}^{t}, \bm{W}, \bm{X})$~~
                \textcolor{gray}{\texttt{\# Calculate token-weighted parameter importance by CIM}}\\

                \If{$i~\%~\mathcal{J}=0$}{
                    $\bm{M}^p\gets \mathbb{I}(\bm{S}'^p> \theta^p)$~~
                    \textcolor{gray}{\texttt{\# Select parameters to prune with $\theta^p\in\{\theta^p_v, \theta^p_l, \theta^p_c\}$}}\\
                    \textcolor{gray}{\texttt{\# Decay the mask to $0$ in $5\mathcal{J}$ steps; the last $\mathcal{J}$ steps perform recovery only}}\\
                    $\bm{M}^p \gets \bm{M}^p + (1-i/(5\mathcal{J}))(1-\bm{M}^p)$\\
                    $\bm{S}'^p\gets 0$~~
                    \textcolor{gray}{\texttt{\# Reset the accumulated parameter importance}}
                }
            }

            $\bm{W}\gets \bm{W} - \nabla_{\bm{W}} \mathcal{L}(\bm{X}^L, Y)$~~
            \textcolor{gray}{\texttt{\# Calculate loss and update parameters}}
        }

        $\textit{LA}\gets \textit{CA}$, $\textit{LF}\gets \textit{CF}$~~
        \textcolor{gray}{\texttt{\# Record accuracy and FLOPs at previous stage}}
        
        $\textit{CA},~\textit{CF}\gets \text{Evaluate}(\mathcal{F}, \mathcal{D}_{\text{val}})$~~
        \textcolor{gray}{\texttt{\# Re-evaluate current accuracy and FLOPs}}

        \textcolor{gray}{\texttt{\# Calculate the pruning cost at current stage}} \\
        $\Delta \mathit{val\_acc} \gets \mathit{LA}-\mathit{CA}$,~$\Delta \mathit{FLOPs}\gets \mathit{LF}-\mathit{CF},~r\gets \Delta \mathit{val\_acc} / \Delta\mathit{FLOPs}$

        $I_m\gets T - \mathcal{T}_m$~~
        \textcolor{gray}{\texttt{\# Get the interval since last execution of $\mathcal{B}_m$}}
        
        $\mathcal{R}_m \gets \text{Get\_cost\_with\_history}(\mathcal{R}_m, r, I_m)$~~
        \textcolor{gray}{\texttt{\# Update pruning cost with~\cref{eq:ema_cost}}}

        $\mathcal{T}_m \gets T$~~
        \textcolor{gray}{\texttt{\# Update the pruning stage where $\mathcal{B}_m$ is last performed}}

        \textcolor{gray}{\texttt{\# Shift mode by random exploration and historical information}}

        \If{$\mathrm{Sample\_from\_uniform}()<\rho$}{
            \textcolor{gray}{\texttt{\# Select a pruning mode randomly according to~\cref{eq:rho_m}}} \\
            $\rho_0, \rho_1, \rho_2, \rho_3, \rho_4 \gets \text{Get\_probability}(T,\mathcal{T})$\\
            $m \gets \text{Random\_choice}(\text{Indices}(\mathcal{B}), \text{prob}=\{\rho_0, \rho_1, \rho_2, \rho_3, \rho_4\})$ 
        }   
        \Else {
            $m \gets \arg \min (\mathcal{R})$~~
            \textcolor{gray}{\texttt{\# Select a priority pruning mode according to pruning cost}}
        }
    }
    $\hat{\bm{W}}\gets \bm{W}\odot{\bm{M}^p}$~~
    \textcolor{gray}{\texttt{\# Completely prune parameters according to the parameter mask}}

    $\hat{\theta_v^t}\gets \Theta_3$, $ \hat{\theta_l^t} \gets \Theta_4$
    
    \Return $\mathcal{F}(\cdot|\hat{\bm{W}}, \hat{\theta_v^t}, \hat{\theta_l^t})$
    
\end{algorithm*}

The overall workflow of our proposed CoMP is summarized in~\cref{psudocode}.
Lines $13 \sim 30$ describe the pipeline of inner loop as shown in~\cref{fig:framework-a}, pruning parameters and tokens based on importance scores calculated by the CIM module. Lines $8\sim 44$ depict the pipeline of outer loop as displayed in~\cref{fig:framework-a}, by periodically shifting pruning modes with random exploration and historical information.
Notably, $\mathcal{I}$ training steps are performed between mode shifting. In the parameter pruning mode, this interval is evenly divided into six sub-intervals $\mathcal{J}\!=\!\mathcal{I}/6$. For each of the first five sub-intervals, we accumulate parameter importance scores over the $\mathcal{J}$ steps to ensure training stability, and then uniformly decay the mask values $\bm{M}^p$ from 1 to 0, by following the UPop \cite{upop} framework. In the final $\mathcal{J}$ steps, only parameter updates are performed to enable model recovery.

\section{Detailed Description of Datasets and Evaluation Metrics}
\label{sec:desc_dataset_metric}
For task-specific VLMs, we primarily evaluate our proposed method on four widely used public datasets, including NLVR2 \cite{nlvr2}, COCO \cite{coco}, Flickr30K \cite{flickr30k} and VQAv2 \cite{vqav2}.
We also include experiments on the MSR-VTT \cite{msrvtt} dataset in this document.
For LLaVA, we further assess methods on six widely adopted image understanding benchmarks, including MME \cite{mme}, MMBench \cite{mmbench}, GQA \cite{gqa}, TextVQA \cite{textvqa}, VQAv2 \cite{vqav2} and POPE \cite{pope}. In addition, the pruning is conducted through supervised fine-tuning (SFT) on the official 665K instruction data \cite{llavaimproved} (denoted as Mix665K). 

\paragraph{NLVR2}
The NLVR2 \cite{nlvr2} dataset is designed for the advancement of joint reasoning over natural language and images. 
It contains $29{,}680$ sentences and $127{,}502$ real-world images, which are combined to form $107{,}292$ examples. 
Each example consists of a caption text paired with two images, and the task is to determine whether the caption accurately describes both images. 
The overall examples are divided into training, development, public test and unreleased test sets, which contain $86{,}373$, $6{,}982$, $6{,}967$ and $6{,}970$ examples, respectively.
We report the accuracy on the development set (Dev. Acc.) and the public test set (Test Acc.) as the evaluation metric.

\paragraph{COCO}
The COCO \cite{coco} dataset is a widely used benchmark that facilitates multiple tasks by promoting the scene understanding capability. Each image is annotated with 5 captions. 
In our work, by following \cite{blip}, we adopt the Karpathy split for both image-text retrieval and image captioning tasks, $\emph{i.e.}$ $113$K, $5$K and $5$K images for training, validation and test, respectively.
In the image-text retrieval task, as the goal is to retrieve the most relevant image/text given an input query, we report the top-1/-5 recall (R@1/5) as the evaluation metric.
In the image captioning task, as this task aims to generate a descriptive sentence that accurately reflects the visual content of a given image, we report CIDEr \cite{cider} and SPICE \cite{spice} as the evaluation metrics. Concretely, 
CIDEr \cite{cider} evaluates the quality of the generated captions by measuring their consensus with multiple reference sentences. SPICE \cite{spice} simulates the human judgment process by comparing the semantic propositional content.

\paragraph{Flickr30K}
The Flickr30K \cite{flickr30k} dataset is originally designed for visual denotation. We also adopt the Karpathy split on the image-text retrieval task and report the top-1/-5 recall (R@1/5) metrics for evaluation.
The training, validation and test sets consist of $29$K, $1$K and $1$K images, respectively, where each image has 5 captions.

\begin{table*}[t]
  \footnotesize
  \centering
  \caption{Training and testing configurations for pruning BLIP, CLIP and LLaVA models on various vision-language tasks. `Reasoning', `Retrieval', `Captioning', `VQA', `Und' denote the visual reasoning task, the image-text retrieval task, the image captioning task, the visual question answering task and the image understanding task, respectively. $\text{R}_{\text{I}\rightarrow \text{T}}$ and $\text{R}_{\text{T}\rightarrow \text{I}}$ denote the recall for image-to-text and text-to-image retrieval, respectively. $-\text{Val\_Loss}$ denotes the negative of the loss on validation set. `-' indicates that the setting is not applied to the corresponding task.}
  \label{tab:hyperparameter_table}
  \begin{tabular}{c @{\hspace{2.0\tabcolsep}} c @{\hspace{1.0\tabcolsep}} c @{\hspace{1.0\tabcolsep}} c @{\hspace{1.0\tabcolsep}} c @{\hspace{1.0\tabcolsep}} c @{\hspace{1.0\tabcolsep}} c @{\hspace{1.0\tabcolsep}} c @{\hspace{1.0\tabcolsep}} c @{\hspace{1.0\tabcolsep}} }
  \toprule[1pt]
  \multirow{2}{*}{Configurations} & BLIP-Reasoning & \multicolumn{2}{c}{BLIP-Retrieval} & BLIP-Captioning & BLIP-VQA & \multicolumn{2}{c}{CLIP-Retrieval} & LLaVA-Und \\ 
    \cmidrule{2-9}
     & NLVR2 \cite{nlvr2} & COCO \cite{coco} & Flickr30K \cite{flickr30k} & COCO \cite{coco} & VQAv2 \cite{vqav2} & COCO \cite{coco} & Flickr30K \cite{flickr30k} & Mix665K \cite{llavaimproved} \\
    \midrule
    \makecell[l]{Train batch size} & 32 & 64 & 16 & 64 & 16 & 16 & 16 & 32 \\
    \makecell[l]{Test batch size} & 64 & 64 & 32 & 64 & 16 & 32 & 32 & 1 \\
    \makecell[l]{Train epochs} & 30 & 10 & 18 & 9 & 2 & 10 & 10 & 1 \\
    \makecell[l]{Learning rate} & 3e-6 & 1e-7 & 1e-7 & 1e-5 & 5e-6 & 5e-6 & 5e-6 & 2e-5 \\
    \makecell[l]{Step size $\mathcal{S}_P$} & 0.02 & 0.02 & 0.02 & 0.02 & - & 0.01 & 0.01 & 0.0005\\
    \makecell[l]{Step size $\mathcal{S}_T$} & 0.4 & 0.4 & 0.4 & 0.2 & - & 0.2 & 0.2 & 0.02/0.01 \\
    \makecell[l]{Interval $\mathcal{I}$} & 300 & 300 & 300 & 300 & - & 300 & 150 & 18 \\
    \makecell[l]{Metric $\mathit{val\_acc}$} & Dev. Acc. & \multicolumn{2}{c}{$(\text{R}_{\text{I}\rightarrow \text{T}}@1 + \text{R}_{\text{T}\rightarrow \text{I}}@1)/2$} & $(\text{SPICE+CIDEr})/2$ & - & \multicolumn{2}{c}{$(\text{R}_{\text{I}\rightarrow \text{T}}@1 + \text{R}_{\text{T}\rightarrow \text{I}}@1)/2$} & $-\text{Val\_Loss}$ \\
    \makecell[l]{Weight Decay} & \multicolumn{5}{c}{0.05} & \multicolumn{2}{c}{0.2} & 0 \\
    \bottomrule[1pt]
  \end{tabular}
\end{table*}

\begin{table*}[t]
  \footnotesize
  \centering
  \caption{Configurations of the models for multiple tasks. \textsuperscript{$*$} indicates two images share one vision transformer in a single forward propagation during inference.}
  \label{tab:config_table}
  \begin{tabular}{c@{\hspace{2.0\tabcolsep}}c@{\hspace{1.25\tabcolsep}}c@{\hspace{1.25\tabcolsep}}c@{\hspace{1.25\tabcolsep}}c@{\hspace{1.25\tabcolsep}}c@{\hspace{1.25\tabcolsep}}c@{\hspace{1.25\tabcolsep}}c@{\hspace{1.25\tabcolsep}}c@{\hspace{1.25\tabcolsep}}c@{\hspace{1.25\tabcolsep}}c@{\hspace{1.25\tabcolsep}}c@{\hspace{1.25\tabcolsep}}}
    \toprule[1pt]
    \multirow{2}{*}{Task} & \multirow{2}{*}{\makecell{Input\\resolution}} & \multicolumn{5}{c}{Vision Transformer} & \multicolumn{5}{c}{Language Transformer} \\
    & & Number & Layers & Width & Heads & Intermediate & Number & Layers & Width & Heads & Intermediate \\
    \midrule
    \makecell[l]{BLIP-Reasoning} & 384$\times$384 & 2\textsuperscript{$*$} & 12 & 768 & 12 & 3072 & 1 & 12 & 768 & 12 & 3072 \\
    \makecell[l]{BLIP-Retrieval} & 384$\times$384 & 1 & 12 & 768 & 12 & 3072 & 1 & 12 & 768 & 12 & 3072 \\
    \makecell[l]{BLIP-Captioning} & 384$\times$384 & 1 & 12 & 768 & 12 & 3072 & 1 & 12 & 768 & 12 & 3072 \\
    \makecell[l]{BLIP-VQA} & 480$\times$480 & 1 & 12 & 768 & 12 & 3072 & 2 & 12 & 768 & 12 & 3072 \\
    \makecell[l]{CLIP-Retrieval} & 336$\times$336 & 1 & 24 & 1024 & 16 & 4096 & 1 & 12 & 768 & 12 & 3072 \\
    \makecell[l]{LLaVA-Image Understanding} & 336$\times$336 & 1 & 24 & 1024 & 16 & 4096 & 1 & 32 & 4096 & 32 & 11008 \\
    \bottomrule[1pt]
  \end{tabular}
\end{table*}

\paragraph{VQAv2}
The VQAv2 \cite{vqav2} dataset is curated for the visual question answering task.
Given an image and a question, this task encourages the model to jointly understand and reason over all information and generate a short answer.
For the BLIP model, by following \cite{blip}, we adopt the split consisting of $123$K images and $658$K questions for training, and $81$K images and $448$K questions for testing. We report accuracy on both the development (Test-dev) and standard (Test-std) splits of the test set.
For the LLaVA model, by following \cite{llavaimproved}, we conduct zero-shot evaluation and report accuracy on the development split of the test set.
All metrics are obtained through the official evaluation website.

\paragraph{MSR-VTT}
The MSR-VTT \cite{msrvtt} dataset is a large-scale video description benchmark with comprehensive categories and varied video content. It contains $10$K video clips, each paired with 20 human-annotated textual descriptions. We follow the evaluation protocol of the BLIP model \cite{blip} to perform zero-shot video-text retrieval and report the top-1/-5 recall (R@1/5) metrics on the $1$K test split.

\begin{figure}[t]
\centering
\includegraphics[width=0.85\linewidth]{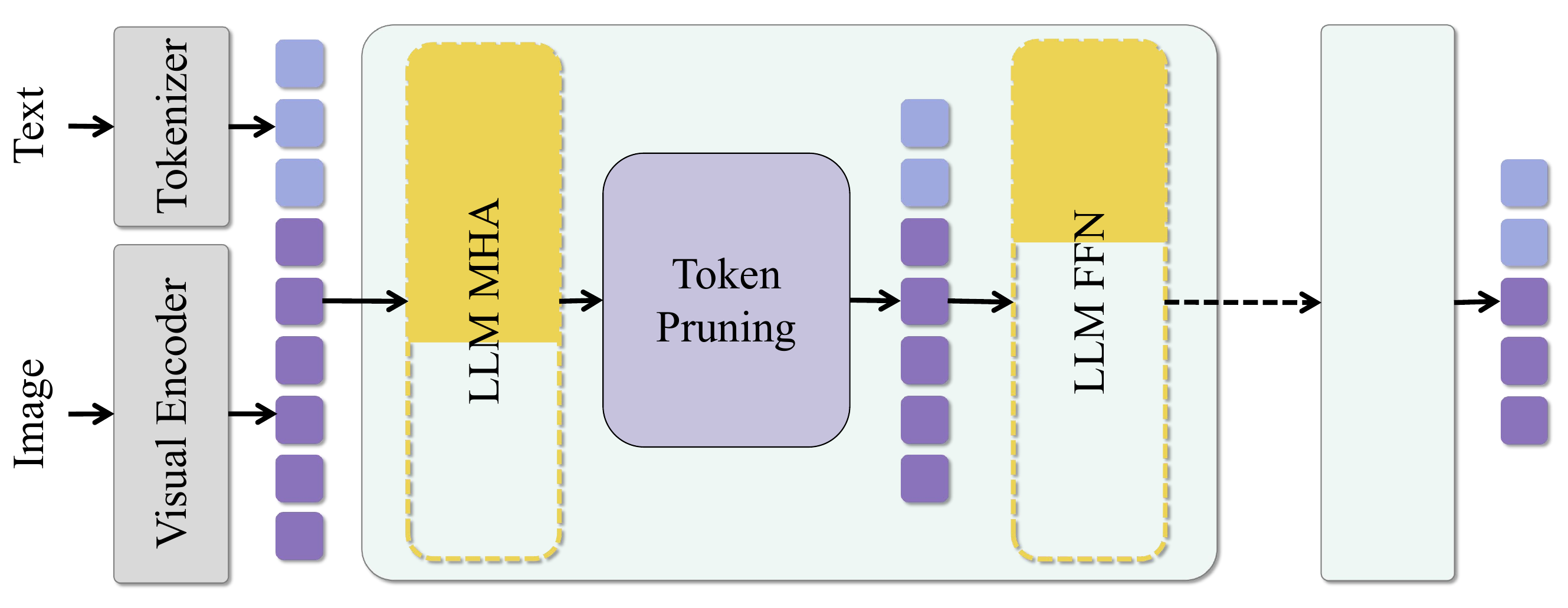}
\caption{Illustration of extending CoMP to the LLaVA-style architecture. We perform pruning in the LLM component, including pruning of vision tokens, language tokens and parameters, since this part accounts for more than 95\% of the overall FLOPs.}
\label{fig:illus_llava}
\end{figure}

\paragraph{MME}
The MME \cite{mme} dataset assesses a model’s perception and cognition capabilities. It consists of 14 sub-tasks with $2{,}000$ questions for perception and 800 for cognition, where each image is associated with two binary (Yes/No) queries. By following existing works \cite{visionzip,dart,pyramiddrop}, we report the summation of scores on both capabilities. For the average score in~\cref{tab:llava_comparison}, the summation is normalized by dividing by $2{,}800$, which corresponds to the full score.

\paragraph{MMBench}
The MMBench~\cite{mmbench} dataset provides a balanced and comprehensive evaluation of model capabilities across three hierarchical levels. It consists of $3{,}217$ multiple-choice questions, with the dataset split into development and test subsets at a $4{:}6$ ratio. By following \cite{llavaimproved}, we report the accuracy on the development set by submitting predictions to the official evaluation server.

\paragraph{GQA}
The GQA \cite{gqa} dataset targets visual understanding and reasoning capabilities in complex real-world scenarios. It contains binary and open queries across real-world reasoning, scene understanding, and compositional question answering. In line with the compared methods, we perform evaluation on the `testdev' subset from \cite{llavaimproved}, which contains $12{,}578$ questions, and report the standard accuracy.

\paragraph{TextVQA}
The TextVQA \cite{textvqa} dataset assesses a model’s ability to understand and reason over text present within images. The benchmark requires the model to read textual content from images and answer associated open-ended questions. We report accuracy on the validation split, which comprises $5{,}000$ image-question pairs.

\begin{figure*}[t]
    \centering
    \includegraphics[width=0.95\textwidth]{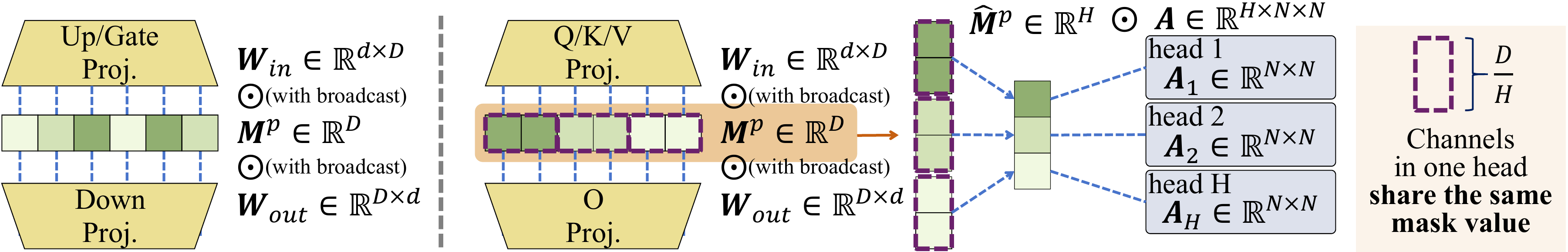}
    \caption{Illustration of the dimension mapping for parameter pruning in FFN (Left) and MHA (Right).}
    \label{fig:dim_map}
\end{figure*}

\begin{figure*}[t]
    \centering
    \includegraphics[width=0.95\textwidth]{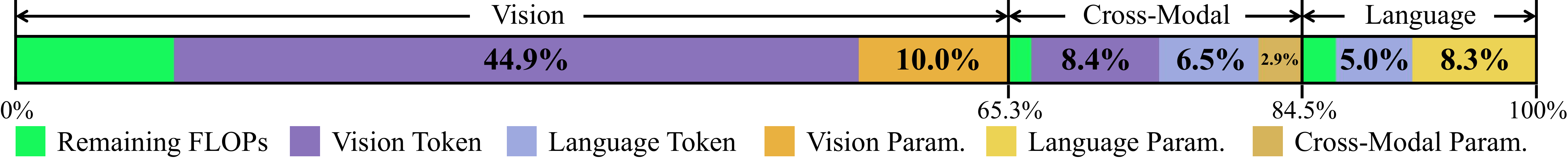}
    \caption{FLOPs contributions of different pruning modes for the BLIP model, evaluated on the NLVR2 dataset at a pruning ratio of 0.85.}
    \label{fig:mode_contribution}
\end{figure*}

\begin{table}[!t]
    \centering
    \caption{Comparison of GFLOPs, Latency (ms per image) and Speedup Ratio across various pruning methods on BLIP using NLVR2 dataset at a pruning ratio of 0.85. Best results are highlighted in \textbf{bold}.}
    \label{tab:inference_latency}
    \footnotesize
    \begin{tabular}{l|ccc}
        \toprule[1pt]
        \makecell[c]{Method} & GFLOPs & Latency $\downarrow$ & Speedup Ratio $\uparrow$ \\
        \midrule
        Uncompressed         & 132.54 & 4.85               & 1.00$\times$  \\
        UPop \cite{upop}     & 20.01  & 2.61               & 1.86$\times$  \\
        MADTP \cite{madtp}   & 20.57  & 1.27               & 3.82$\times$  \\
        \textbf{CoMP (Ours)} & 20.26  & \textbf{1.25}      & \textbf{3.85$\times$} \\
        \bottomrule[1pt]
    \end{tabular}
\end{table}

\paragraph{POPE}
The POPE \cite{pope} dataset evaluates the object hallucination of models. It is built upon $500$ images selected from the COCO \cite{coco} dataset. For each image, a polling-based query with 6 questions is performed, organized into three subsets based on different sampling strategies: random, common, and adversarial. By following \cite{llavaimproved}, we report the average F1 score across all three subsets.

\paragraph{Mix665K}
The Mix665K \cite{llavaimproved} data is a mixed collection derived from multiple benchmarks to comprehensively enhance the model’s capabilities. It comprises $665$K image–conversation pairs and serves as the official training set for LLaVA’s supervised fine-tuning process.

\section{Detailed Implementation Details}
\label{sec:detailed_implementation}
To evaluate the effectiveness of our method, we compress the BLIP \cite{blip} and CLIP \cite{clip} models fine-tuned for multiple downstream tasks, by following \cite{upop} and \cite{madtp}. 
We further extend our methods to the LLaVA-v1.5-7B model.
The training and testing configurations are summarized in~\cref{tab:hyperparameter_table}.
We also report the GFLOPs/TFLOPs of the model during inference, of which the corresponding configurations are summarized in~\cref{tab:config_table}.

Our collaborative pruning method builds upon the existing parameter and token pruning frameworks \cite{upop,madtp,pyramiddrop}, and the multi-mode progressive pruning is achieved by adjusting the corresponding hyperparameters (\emph{i.e.} to generate the $\Delta\Theta_m$ in~\cref{psudocode}).
Specifically, we utilize the parameter pruning framework from UPop \cite{upop} and, at each stage, increase its parameter reduction ratio by a step size of $\mathcal{S}_P$ for parameter pruning.
Meanwhile for token pruning, we adopt the MADTP \cite{madtp} framework for BLIP/CLIP, and progressively adjust the temperature it defines via a step size of $\mathcal{S}_T$. Notably, since MADTP is not suitable for LLaVA, we instead use the PDrop \cite{pyramiddrop} framework, adjusting its token ratio hyperparameter for pruning. Given that language tokens are far fewer than vision tokens, we set $\mathcal{S}_T$ to 0.02/0.01 for vision/language tokens, respectively, as shown in~\cref{tab:hyperparameter_table}.
Pruning is conducted simultaneously with model optimization. Throughout all `Train epochs', we apply our proposed CoMP method to progressively prune the model until reaching the target FLOPs, after which we fix the pruning configuration and fine-tune the pruned model to recover its performance.

\begin{table}[!t]
    \centering
    \caption{Ablation results of the random exploration ratio $\rho$ on BLIP using NLVR2 dataset at a pruning ratio of 0.8. The default setting is \underline{underlined}.}
    \label{tab:ablation_rho}
    \footnotesize
    \begin{tabular}{c | c c| c}
    \toprule[1pt]
    $\rho$ & Dev. Acc. & Test Acc. & GFLOPs \\
    \midrule
    0.1 & 78.70 & 79.08 & 26.00 \\
    \underline{0.2} & 79.23 & 79.62 & 25.97 \\
    0.3 & 79.50 & 79.76 & 25.95 \\
    0.4 & 78.79 & 79.55 & 26.68 \\
    \bottomrule[1pt]
    \end{tabular}
\end{table}

In addition, we set $\rho = 0.2$, $\tau = 5$ in~\cref{eq:rho_m} for random exploration, and $\lambda_0=0.4$, $I_{\text{max}}=5$ for historical information in the MPS module. These hyperparameters are fixed across all experiments, and ablation studies are provided in~\cref{sec:hyperparameter_ablation}.

For BLIP/CLIP, all experiments are conducted on 2 NVIDIA A800 GPUs. By following the training settings of \cite{blip} and \cite{clip}, we adopt the AdamW \cite{adamw} optimizer, along with a cosine learning rate scheduler \cite{cos_lr} and random data augmentation \cite{randomaug}. We adopt the task-specific evaluation metric on the validation subset to compute $\mathit{val\_acc}$, as defined in~\cref{eq:prune_cost}. For tasks with multiple metrics, we utilize a simple average of the primary metrics, with details provided in~\cref{tab:hyperparameter_table}. Alternatively, max-normalization may be preferable for metrics with varying scales. For LLaVA, given its architectural characteristics, we collaboratively prune the vision tokens, language tokens and parameters in the LLM component, as presented in~\cref{fig:illus_llava}.
A one-epoch SFT is performed on 8 NVIDIA A800 GPUs, with all settings consistent with the uncompressed model \cite{llavaimproved}. Due to the abscence of validation set, we uniformly sample 3K examples from the full 665K training set for validation and adopt the negative loss (denoted as `$-\text{Val\_Loss}$') as a proxy for $\mathit{val\_acc}$.

\begin{table}[!t]
    \centering
    \caption{Ablation results of the decay factor $\lambda$ (parameterized by $\lambda_0$ and $I_{\text{max}}$) on BLIP using NLVR2 dataset at a pruning ratio of 0.8. The default setting is \underline{underlined}.}
    \label{tab:ablation_lambda}
    \footnotesize
    \begin{tabular}{@{\hspace{2pt}} c @{\hspace{3pt}} c @{\hspace{3pt}} l | c @{\hspace{7pt}} c | @{\hspace{2pt}} c @{\hspace{2pt}}}
    \toprule[1pt]
    \multicolumn{3}{c|}{Setting} & \multirow{2}{*}{\makecell{Dev.\\ Acc.}} & \multirow{2}{*}{\makecell{Test \\ Acc.}} & \multirow{2}{*}{GFLOPs} \\
    $\lambda_0$ & $I_{\text{max}}$ & ~Formulation of $\lambda$ &  &  &  \\
    \midrule
    0.3 & 1 & $\lambda\!=\!0.3$ & 79.07 & 79.03 & 26.51 \\
    0.4 & 1 & $\lambda\!=\!0.4$ & 79.18 & 79.33 & 25.93 \\
    0.5 & 1 & $\lambda\!=\!0.5$ & 78.75 & 79.02 & 25.91 \\
    \midrule
    0.4 & 2 & $\lambda\!=\!\max(0.4\!-\!0.20(I_m\!-\!1), 0)$ & 78.77 & 78.79 & 26.32 \\
    0.4 & 4 & $\lambda\!=\!\max(0.4\!-\!0.10(I_m\!-\!1), 0)$ & 79.35 & 79.45 & 26.31 \\
    0.4 & 5 & \underline{$\lambda\!=\!\max(0.4\!-\!0.08(I_m\!-\!1), 0)$} & 79.23 & 79.62 & 25.97 \\
    0.4 & 6 & $\lambda\!=\!\max(0.4\!-\!0.07(I_m\!-\!1), 0)$ & 79.25 & 79.61 & 26.19 \\
    0.4 & 8 & $\lambda\!=\!\max(0.4\!-\!0.05(I_m\!-\!1), 0)$ & 79.12 & 79.45 & 26.33 \\
    \bottomrule[1pt]
    \end{tabular}
\end{table}

\section{Dimension Mapping for Parameter Pruning}
\label{sec:dim_map}
\cref{fig:dim_map} illustrates the detailed dimension mapping process in parameter pruning. In the FFN module, all operations follow the standard per-channel definition described in~\cref{sec:methodology}. 
For the MHA module, pruning is performed in a head-wise manner, \ie, all channels within a head are pruned simultaneously to preserve parallelism after compression. Taking the $h$-th head as an example, this requires that the mask values within the head remain identical, which can be formally expressed as:
\begin{equation}
    \bm{M}^p_{(h-1)d_k+1}=\bm{M}^p_{(h-1)d_k+2}=...=\bm{M}^p_{hd_k}.
\end{equation}
Meanwhile, the prunable units are reduced from $D$ channels to $H$ heads. For each unit, its importance score is calculated as the average of the channel-level importance scores within the corresponding head:
\begin{equation}
    \bm{S}'^p_h = \frac{1}{d_k}\sum_{i=(h-1)d_k+1}^{hd_k} \bm{S}'^p_{i,:}.
\end{equation}
Furthermore, in our CIM module, the mask $\bm{M}^p$ is transferred from the channel dimension to the attention head dimension, resulting in $\hat{\bm{M}}^p$ as shown in~\cref{eq:mask_attention}, where $\hat{\bm{M}}^p_h=\bm{M}^p_{hd_k}$.

\section{More Experimental Results and Analysis}
\label{sec:more_experiments}

\subsection{Real-World Inference Latency}
\label{sec:inference_latency}
CoMP employs a structured pruning scheme, which is widely recognized for its ease of model deployment without requiring hardware-specific adaptations \cite{upop, samp}. 
To demonstrate the real-world efficiency of CoMP, we report the inference latency and speedup ratio on a single RTX 4090 GPU in~\cref{tab:inference_latency}. On BLIP-NLVR2 with a FLOPs pruning ratio of 0.85, CoMP achieves a 3.85$\times$ speedup while maintaining state-of-the-art accuracy, as reported in~\cref{tab:sota_table}. 

\begin{table}[!t]
    \centering
    \caption{Ablation results of the interval steps $\mathcal{I}$ on BLIP using the NLVR2 dataset at a pruning ratio of 0.8. The default setting is \underline{underlined}.}
    \label{tab:ablation_I}
    \footnotesize
    \begin{tabular}{c | c c| c}
        \toprule[1pt]
        $\mathcal{I}$ & Dev. Acc. & Test Acc. & GFLOPs \\
        \midrule
        240 & 79.42 & 79.25 & 26.50 \\
        \underline{300} & 79.23 & 79.62 & 25.97 \\
        360 & 79.09 & 79.71 & 26.19 \\
        \bottomrule[1pt]
    \end{tabular}
\end{table}

\begin{table}[!t]
    \centering
    \caption{Ablation results of $\tau$ in ~\cref{eq:rho_m} on BLIP using the NLVR2 dataset at a pruning ratio of 0.8. The default setting is \underline{underlined}.}
    \label{tab:ablation_tau}
    \footnotesize
    \begin{tabular}{c | c c| c}
        \toprule[1pt]
        $\tau$ & Dev. Acc. & Test Acc. & GFLOPs \\
        \midrule
        1 & 79.30 & 79.69 & 26.38 \\
        \underline{5} & 79.23 & 79.62 & 25.97 \\
        10 & 79.12 & 79.49 & 26.68 \\
        \bottomrule[1pt]
    \end{tabular}
\end{table}

\subsection{Effect of Distinct Pruning Modes}
\label{sec:mode_contribution}
To demonstrate the overall effect of collaborative pruning, \cref{fig:mode_contribution} presents the FLOPs reduction contributed by the five pruning modes on BLIP-NLVR2 at a pruning ratio of 0.85. 
For instance, in the full model (\ie 100\% FLOPs), the vision branch accounts for 65.3\% of the total computation, where vision parameter pruning and vision token pruning contribute FLOPs reductions of 10.0\% and 44.9\%, respectively.

\subsection{Additional Ablations Results}
\label{sec:hyperparameter_ablation}

\paragraph{On Random Exploration Ratio $\rho$}
We ablate the random exploration ratio $\rho$ in the MPS module on BLIP using the NLVR2 dataset at a pruning ratio of 0.8. 
As~\cref{tab:ablation_rho} shows, the best performance is achieved when $\rho$ is set between 0.2 and 0.3. Intuitively, a small $\rho$ constrains the model's exploration capability, potentially leading to stuck in sub-optimal pruning modes. Conversely, an excessively large $\rho$ may degrade the overall performance with unstable pruning process. The default value of 0.2 represents a balanced trade-off and consistently yields competitive results.

\begin{table*}[!t]
    \centering
    \caption{Comparison of Dev./Test Acc. (\%), R@1/5 (\%) and GFLOPs by CoMP and MADTP on BLIP for NLVR2 visual reasoning and COCO image-text retrieval tasks. Reported as mean $\pm$ std over 5 seeds. The best results are highlighted in \textbf{bold}.}
    \label{tab:stability}
    \footnotesize
    \begin{tabular}{@{\hspace{1.0\tabcolsep}} c @{\hspace{1.0\tabcolsep}}| c @{\hspace{1.0\tabcolsep}} c | c | c @{\hspace{1.0\tabcolsep}} c | c @{\hspace{1.0\tabcolsep}} c | c }
    \toprule[1pt]
    \multirow{3}{*}{Method} & \multicolumn{3}{c|}{NLVR2-Reasoning} & \multicolumn{5}{c}{COCO-Retrieval} \\
    \cline{2-9}
                            & \multirow{2}{*}{Dev. Acc.} & \multirow{2}{*}{Test Acc.} & \multirow{2}{*}{GFLOPs} & \multicolumn{2}{c|}{I$\rightarrow$T} & \multicolumn{2}{c|}{T$\rightarrow$I} & \multirow{2}{*}{GFLOPs} \\
      & & & & R@1 & R@5 & R@1 & R@5 & \\
    \midrule
    \makecell[l]{MADTP \cite{madtp}} & 77.16$\pm$0.57 & 77.64$\pm$0.47 & 26.77$\pm$0.23 & 74.38$\pm$0.32 & 91.50$\pm$0.24 & 56.28$\pm$0.33 & 80.74$\pm$0.13 & 30.04$\pm$0.44 \\
    \makecell[l]{\textbf{CoMP (Ours)}} & \textbf{79.13}$\pm$0.24 & \textbf{79.67}$\pm$0.28 & 26.33$\pm$0.38 & \textbf{75.96}$\pm$0.27 & \textbf{92.54}$\pm$0.13 & \textbf{57.32}$\pm$0.31 & \textbf{81.44}$\pm$0.30 & 29.24$\pm$0.32 \\
    \bottomrule[1pt]
    \end{tabular}
\end{table*}

\begin{table}[!t]
    \centering
    \caption{Comparison of R@1/5 (\%) and GFLOPs for BLIP on MSR-VTT dataset for the zero-shot video-text retrieval task. The best results are highlighted in \textbf{bold}.}
    \label{tab:comparison_blip_msr_vtt}
    \footnotesize
    \begin{tabular}{@{\hspace{2pt}} c @{\hspace{2pt}}|@{\hspace{2pt}} c @{\hspace{2pt}}|@{\hspace{2pt}} c @{\hspace{2pt}}|@{\hspace{3pt}} c @{\hspace{6pt}} c @{\hspace{3pt}}|@{\hspace{3pt}} c @{\hspace{6pt}} c @{\hspace{3pt}}|@{\hspace{2pt}} c @{\hspace{2pt}}}
    \toprule[1pt]
    \multirow{2}{*}{Method} & \multirow{2}{*}{\makecell{Pruning \\ Mode}} & \multirow{2}{*}{\makecell{Pruning \\ Ratio}} & \multicolumn{2}{@{\hspace{3pt}}c@{\hspace{3pt}}|@{\hspace{3pt}}}{V$\rightarrow$T} & \multicolumn{2}{@{\hspace{3pt}}c@{\hspace{3pt}}|@{\hspace{2pt}}}{T$\rightarrow$V} & \multirow{2}{*}{GFLOPs} \\
        &  & & R@1 & R@5 & R@1 & R@5 &\\
    \midrule
    \makecell[l]{Uncompressed} & / & / & 35.8 & 59.7 & 43.3 & 65.5 & 733.4 \\
    \makecell[l]{UPop \cite{upop}} & P & 0.7 & 19.1 & 37.3 & 24.0 & 42.2 & 256.2 \\
    \makecell[l]{MADTP \cite{madtp}} & T & 0.7 & 34.8 & 59.8 & 38.9 & 62.4 & 265.3 \\
    \makecell[l]{\textbf{CoMP (Ours)}} & C & 0.7 & \textbf{35.4} & \textbf{60.3} & \textbf{39.5} & \textbf{63.1} & 261.1 \\
    \bottomrule[1pt]
    \end{tabular}
\end{table}

\begin{figure}[t]
\centering
\includegraphics[width=0.9\linewidth]{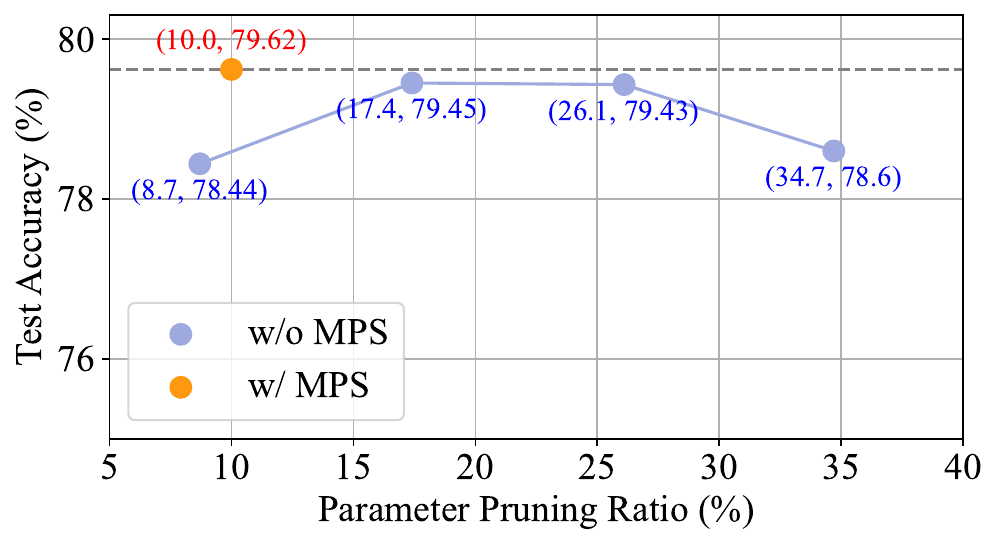}
\caption{Comparison of test accuracy with (w/) and without (w/o) the proposed MPS module on NLVR2 at an overall FLOPs pruning ratio of 0.8.}
\label{fig:ratios_compare}
\end{figure}

\paragraph{On Decay Factor $\lambda$}
The design of $\lambda$ in~\cref{eq:ema_cost} aims to leverage historical information while appropriately discounting the outdated contributions.
That means, if a pruning mode has not been performed for a long time, its past cost information should be diminished when guiding subsequent mode selection.
To validate this, we conduct a two-step ablation study using BLIP on the NLVR2 dataset, as shown in~\cref{tab:ablation_lambda}.
Specifically, we first fix $\lambda$ as a constant to introduce static historical weighting, where $\lambda=0.4$ yields the best performance, and this value is adopted as the initial weight $\lambda_0$ in the decaying formulation.
Next, we incorporate the linear decay mechanism as in~\cref{eq:ema_cost}, where historical information beyond $I_{\text{max}}$ stages is no longer considered.
Empirically, setting $I_{\text{max}}=5$ provides a favorable trade-off.
These results suggest that incorporating historical information can stabilize the pruning process, but its effect should be limited to a reasonable temporal window, thus avoiding reliance on excessively old data.

\paragraph{On Interval Steps $\mathcal{I}$}
The interval $\mathcal{I}$ between mode shifting directly determines the duration of the pruning process. As shown in~\cref{tab:ablation_I}, our method is relatively insensitive to this hyperparameter. In practice, we set $\mathcal{I}$ empirically according to the model size, ensuring that the cost of pruning process does not exceed that of fine-tuning while still enabling sufficient recovery after conducting each pruning mode.

\paragraph{On Softmax Temperature $\tau$}
In~\cref{eq:rho_m}, the temperature $\tau$ modulates the distribution over the stage intervals of modes since their last execution. As reported in~\cref{tab:ablation_tau}, our method exhibits low sensitivity to this . Accordingly, we fix $\tau=5$ for all experiments in this work.

\paragraph{On Additional Effect of MPS}
It's worth noting that the proposed MPS module not only facilitates shifting between pruning modes but also enables adaptively adjusting pruning ratios for parameter and token pruning, under a fixed overall budget of FLOPs. To further evaluate its effectiveness, we compare MPS to the baseline that  manually adopts fixed parameter pruning ratios. As illustrated in~\cref{fig:ratios_compare}, within the same overall budget of FLOPs, MPS consistently outperforms the counterparts adopting distinct fixed pruning ratios, indicating its superior capability in adaptively adjusting pruning ratios for distinct pruning modes. 
The results also imply that in the context of multi-mode pruning, the ultimate performance is influenced not only by the global allocation of parameter and token pruning ratios, but also by the execution order of distinct pruning modes, for which our method provides a promising solution.

\begin{table}[!t]
    \centering
    \caption{Comparison of Dev./Test Acc. (\%) and GFLOPs by CoMP with different parameter pruning frameworks on BLIP using NLVR2 dataset. The best results are highlighted in \textbf{bold}.}
    \label{tab:param_with_iso}
    \footnotesize
    \begin{tabular}{@{\hspace{3pt}}c@{\hspace{3pt}}|@{\hspace{3pt}} c@{\hspace{3pt}}|@{\hspace{3pt}} c @{\hspace{7pt}}c@{\hspace{3pt}}|@{\hspace{3pt}} c@{\hspace{3pt}}}
    \toprule[1pt]
    \multirow{2}{*}{Method} & \multirow{2}{*}{\makecell{Pruning \\ Ratio}} & \multirow{2}{*}{\makecell{Dev. \\ Acc.}} & \multirow{2}{*}{\makecell{Test \\ Acc.}} & \multirow{2}{*}{GFLOPs} \\
    & & & & \\
    \midrule
    \makecell[l]{CoMP w/ UPop \cite{upop}} & 0.8 & \textbf{79.23} & \textbf{79.62} & 25.97 \\
    \makecell[l]{CoMP w/ Isomorphic Pruning \cite{isomorphic}} & 0.8 & 79.09 & \textbf{79.62} & 26.53 \\
    \midrule
    \makecell[l]{CoMP w/ UPop \cite{upop}} & 0.85 & 75.81 & 76.08 & 20.26 \\
    \makecell[l]{CoMP w/ Isomorphic Pruning \cite{isomorphic}} & 0.85 & \textbf{77.23} & \textbf{77.75} & 19.83 \\
    \bottomrule[1pt]
    \end{tabular}
\end{table}

\begin{figure*}[t]
\centering
\includegraphics[width=\textwidth]{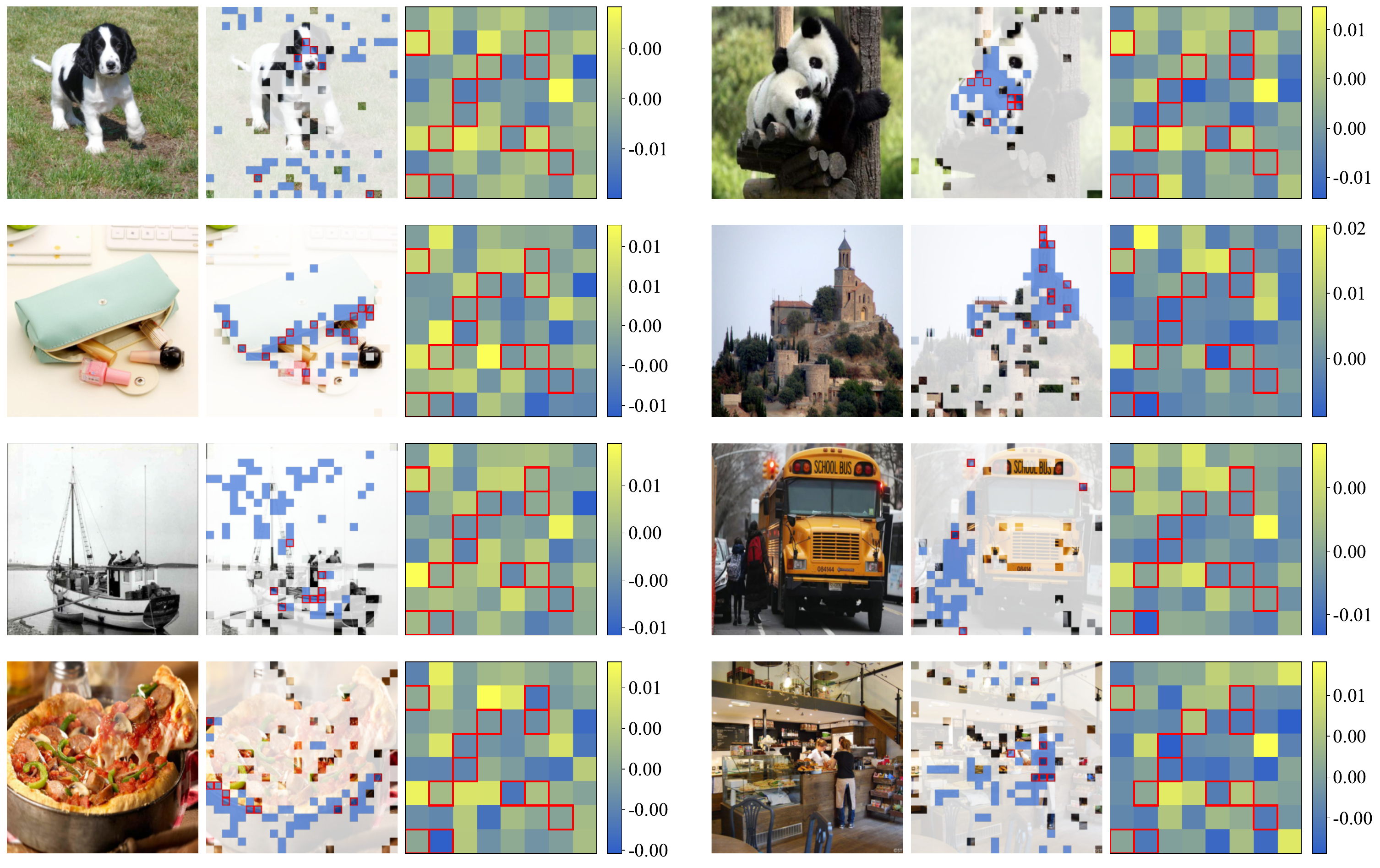}
\caption{Visualization of more examples illustrating inconsistency between parameter importance and token importance in the visual encoder of BLIP. Each example presents, from left to right: (1) the original image; (2) the remaining tokens after token pruning at $\mathit{layer}_{10}$, along with the tokens that contribute most to parameter importance (blue regions), where the overlapping tokens are indicated by red boxes; (3) the heatmap of parameter contributions to token importance at $\mathit{layer}_2$, with the least important parameters (to be pruned by parameter pruning) highlighted by red boxes.}
\label{fig:more_observation}
\end{figure*}

\subsection{Statistical Significance}
\label{sec:more_seeds}
We evaluate the stability and statistical significance of our method using the BLIP model on the visual reasoning and image-text retrieval tasks.
Specifically, we compare our proposed CoMP with the second-best baseline MADTP \cite{madtp}, where each method is evaluated over 5 runs with different random seeds. 
As shown in~\cref{tab:stability}, CoMP consistently achieves higher average performance (\eg~ an average gain of 2.03\% and 1.04\% in NLVR2 Test Acc. and COCO image R@1, respectively) while exhibiting lower standard deviation across most metrics.
Furthermore, paired significance tests indicate these gains are statistically significant ($p=0.0004, 0.0025, 0.0076$ for NLVR2 Test Acc., COCO text R@1 and COCO image R@1, respectively).

\subsection{Extended Evaluation on Video-based Tasks}
\label{sec:evaluation_video}
We further validate the effectiveness of our method on video-based vision-language tasks.
By following the settings of \cite{blip}, we employ the BLIP models compressed on the COCO dataset, as reported in~\cref{tab:other_tasks_table}, to perform zero-shot video-text retrieval on the MSR-VTT \cite{msrvtt} dataset.
As shown in~\cref{tab:comparison_blip_msr_vtt}, our CoMP still consistently outperforms the compared methods, demonstrating the effectiveness and generalization ability of our method on video benchmarks.

\subsection{Orthogonality to Single-Mode Pruning}
\label{sec:orthogonality_param}
Notably, CoMP adaptively schedules multiple pruning modes to progressively achieve the target pruning ratio, optimizing the pruning process through cross-mode collaboration. 
This makes it orthogonal to existing importance criteria-based single-mode pruning frameworks.
Experiments in~\cref{tab:sota_table} and~\cref{tab:llava_comparison} have shown its compatibility with token pruning frameworks like MADTP \cite{madtp} and PDrop \cite{pyramiddrop}.
While for parameter pruning, we further incorporate Isomorphic Pruning \cite{isomorphic}, which is originally developed for vision models, into CoMP by adapting it to the vision-language setting.
We evaluate this configuration on BLIP model using NLVR2 dataset, with all hyperparameters unchanged.
As shown in~\cref{tab:param_with_iso}, at a pruning ratio of 0.8, CoMP combined with either parameter pruning framework achieves similar test accuracy (79.62\%).
However, at a higher ratio of 0.85, integrating Isomorphic Pruning leads to a notable accuracy gain of 1.67\%, owing to its advantage in preserving structural uniformity and mitigating over-pruning.
These results demonstrate the generalizability and flexibility of our CoMP framework, and underscores its strong potential to further reduce accuracy loss by combining with more advanced single-mode pruning techniques.

\subsection{Discussion on Computational Overhead}
\label{sec:computational_overhead}
The CoMP framework introduces additional computational overhead, primarily from the MPS module. Concretely, the mode shifting at each pruning stage requires to re-evaluate model performance on a validation set to compute the $\Delta \mathit{val\_acc}$ term in~\cref{eq:prune_cost}, incurring extra time cost. To reduce this cost, we adopt a reduced validation set: for small datasets like NLVR2 \cite{nlvr2}, we keep the full official validation set, while for the larger datasets like COCO \cite{coco}, we uniformly sample a subset from the official validation set and fix it. As a result, each re-evaluation tasks around 1 minute, which is minor compared to the overall pruning and fine-tuning time.
For instance, at a pruning ratio of 0.8, MADTP \cite{madtp}, UPop \cite{upop} and CoMP take $5.8$h, $12.3$h and $6.7$h on two GPUs, respectively, on the BLIP-NLVR2. We consider this computational overhead generally affordable and acceptable, especially in light of the 2.01\% and 12.13\% improvements in test accuracy that CoMP achieves over these two counterparts.

\subsection{More Examples of Importance Inconsistency}
\label{sec:more_observation}

To further validate the observations introduced in~\cref{sec:introduction}, we provide additional visualizations illustrating the inconsistency between parameter importance metric and token importance metric. Concretely, we adopt the single-mode importance metrics from UPop \cite{upop} and MADTP \cite{madtp}.
For brevity in illustrating parameter behavior, we restrict the visualization to the grouped parameters in MHA, which serve as a representative subset.
As shown in the middle column of each example in~\cref{fig:more_observation}, the tokens identified as important by token pruning rarely coincide with those that contribute most to parameter importance, exhibiting less than 30\% overlap (\ie~highlighted by red boxes). Meanwhile, the right column of each example further demonstrates that parameters considered unimportant by parameter pruning (\ie~highlighted by red boxes) still exert substantial influence on the computation of token importance, as evidenced by high values in the heatmap. These findings indicate that the inconsistency between the two types of importance is widespread, for which our method, especially the CIM module, provides an effective solution for mitigating such cross-mode interference.

\end{document}